\documentclass{article} %
\usepackage{anvil_preprint,times}

\usepackage{amsmath,amsfonts,bm}

\def\eqref#1{equation~\ref{#1}}

\def\1{\bm{1}}
\newcommand{\train}{\mathcal{D}}

\DeclareMathAlphabet{\mathsfit}{\encodingdefault}{\sfdefault}{m}{sl}
\SetMathAlphabet{\mathsfit}{bold}{\encodingdefault}{\sfdefault}{bx}{n}

\usepackage[T1]{fontenc}
\usepackage{microtype}
\usepackage{inconsolata}
\usepackage{graphicx}
\usepackage{booktabs}
\usepackage{hyperref}
\usepackage{url}
\usepackage{amsfonts}
\usepackage{dsfont}
\usepackage{nicefrac}
\usepackage{xcolor}
\usepackage{fix-cm}
\usepackage{longtable}
\usepackage{amsmath}
\usepackage{amssymb}
\usepackage{mathtools}
\usepackage{amsthm}
\usepackage{xspace}
\usepackage{makecell}
\usepackage{enumitem}
\usepackage[capitalize,noabbrev]{cleveref}
\usepackage{subcaption}
\usepackage{algorithm}
\usepackage{algpseudocode}
\usepackage{multirow}
\usepackage{tabularx}
\usepackage{array}
\usepackage{wrapfig}
\usepackage{colortbl}
\usepackage{float}
\usepackage{placeins}
\usepackage{listings}
\usepackage[normalem]{ulem}
\usepackage[most]{tcolorbox}
\usepackage{pifont}
\usepackage{siunitx}
\usepackage{diagbox}
\usepackage{tikz}
\usepackage{fontawesome5}
\usepackage{accsupp}

\usetikzlibrary{arrows.meta,positioning,fit}
\hypersetup{hidelinks}
\hypersetup{
  pdftitle={Anvil: Scaling Embodied Spatial Intelligence through Knowledge Recirculation},
  pdfauthor={Weilin Ruan; Bohan Zhou; Jinghao Li; Yangkai Wei; Wenqian Li; Peiran Dong; Yingjie Xu; Leyi Wu; Haojian Huang; Binghui Xie; Han Yang; Kaiwen Zhou; Yinchuan Li; Ying-Cong Chen; James Cheng},
  pdfsubject={Embodied spatial intelligence; author preprint}
}
\def\method{\textsc{Anvil}\xspace}
\def\bench{\textsc{Anvil-Ben}\xspace}
\def\train{\textsc{Anvil-Train}\xspace}
\def\eval{\textsc{Anvil-Eval}\xspace}
\def\engine{\textsc{Anvil-Gen}\xspace}
\def\model{\textsc{Anvil-MOPD}\xspace} %
\def\basemodel{\textsc{Qwen3.5-9B}\xspace}
\def\benchg{\textsc{Anvil-Bench-G}\xspace}

\definecolor{groundblue}{HTML}{0072B2}
\definecolor{spatialteal}{HTML}{009E73}
\definecolor{actionorange}{HTML}{D55E00}
\definecolor{stressamber}{HTML}{E69F00}
\definecolor{estimateyellow}{HTML}{9B7800}
\definecolor{lossred}{HTML}{A33A3A}
\definecolor{softgray}{HTML}{F2F4F5}
\definecolor{midgray}{HTML}{6B7075}
\definecolor{groundwash}{HTML}{E8F3F9}
\definecolor{spatialwash}{HTML}{E8F5F0}
\definecolor{actionwash}{HTML}{FAEEE9}
\definecolor{losswash}{HTML}{F8ECEC}
\definecolor{titleblue}{HTML}{1369BC}
\definecolor{titlepurple}{HTML}{7953C6}
\definecolor{abstractwash}{HTML}{F2F6FC}

\tcbuselibrary{skins}

\newtcolorbox{introquestions}{
    enhanced,
    frame hidden,
    colback=blue!5,
    borderline west={2pt}{0pt}{blue!70!black},
    sharp corners,
    boxsep=0pt,
    left=8pt,
    right=8pt,
    top=6pt,
    bottom=4pt,
    before skip=7pt plus 2pt minus 1pt,
    after skip=10pt plus 2pt minus 2pt,
}

\newif\ifauthorcolors
\authorcolorsfalse

\ifauthorcolors
  \definecolor{authorblue}{HTML}{185ADB}
  \definecolor{checkorange}{HTML}{E07B00}
\else
  \colorlet{authorblue}{black}
  \colorlet{checkorange}{black}
\fi

\newcommand{\needcheck}[1]{\textcolor{checkorange}{#1}}

\graphicspath{{figs/}{figures/}{tables/}}

\title{Anvil: Scaling Embodied Spatial Intelligence through Knowledge Recirculation}

\newsavebox{\anviltitlebox}
\newcommand{\gradientAnvil}{%
  \sbox{\anviltitlebox}{{\fontsize{19}{22}\selectfont\bfseries Anvil}}%
  \BeginAccSupp{method=escape,ActualText=Anvil}%
  \raisebox{-\dp\anviltitlebox}{\includegraphics[width=\wd\anviltitlebox]{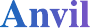}}%
  \EndAccSupp{}%
}

\author{\begin{minipage}{\textwidth}
\centering\normalfont\fontsize{9.5}{13}\selectfont
\textbf{Weilin Ruan}\textsuperscript{1,2},\enspace
\textbf{Bohan Zhou}\textsuperscript{1,2},\enspace
\textbf{Jinghao Li}\textsuperscript{1,2},\enspace
\textbf{Yangkai Wei}\textsuperscript{2},\enspace
\textbf{Wenqian Li}\textsuperscript{2},\\[1pt]
\textbf{Peiran Dong}\textsuperscript{2},\enspace
\textbf{Yingjie Xu}\textsuperscript{2,3},\enspace
\textbf{Leyi Wu}\textsuperscript{2,3},\enspace
\textbf{Haojian Huang}\textsuperscript{2,3},\enspace
\textbf{Binghui Xie}\textsuperscript{2},\\[1pt]
\textbf{Han Yang}\textsuperscript{2},\enspace
\textbf{Kaiwen Zhou}\textsuperscript{2},\enspace
\textbf{Yinchuan Li}\textsuperscript{2},\enspace
\textbf{Ying-Cong Chen}\textsuperscript{3},\enspace
\textbf{James Cheng}\textsuperscript{1,\ensuremath{\dagger}}\\[7pt]
{\fontsize{8}{10.5}\selectfont\color{midgray}
\textsuperscript{1}The Chinese University of Hong Kong\qquad
\textsuperscript{2}Knowin.AI\\[1pt]
\textsuperscript{3}The Hong Kong University of Science and Technology (Guangzhou)\\[3pt]
\textsuperscript{\ensuremath{\dagger}}Corresponding author.\quad
\href{mailto:rwlinno@gmail.com}{\textcolor{groundblue!65!black}{\sffamily\faEnvelope[regular]\enspace\textbf{CONTACT:}\enspace\texttt{rwlinno@}\raisebox{0pt}[0pt][0pt]{\tcbox[on line,colback=groundwash,colframe=groundwash,boxrule=0pt,arc=1pt,boxsep=0pt,left=3pt,right=3pt,top=1pt,bottom=1pt]{\color{groundblue!65!black}\texttt{gmail.com}}}}}}\\[8pt]
\makebox[124pt][c]{\raisebox{-0.5\height}{\includegraphics[width=124pt]{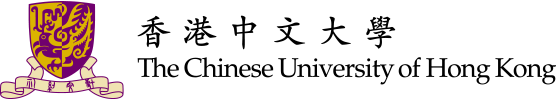}}}\hspace{10pt}%
\makebox[124pt][c]{\raisebox{-0.5\height}{\includegraphics[width=80pt]{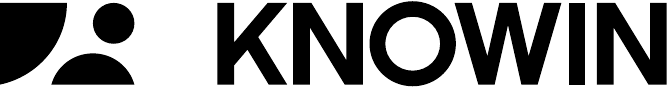}}}\hspace{10pt}%
\makebox[124pt][c]{\raisebox{-0.5\height}{\includegraphics[width=117pt]{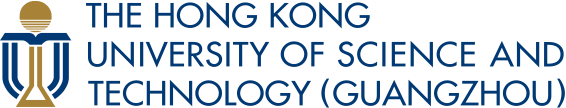}}}\\[12pt]
{\small\sffamily
\href{https://knowin-29.github.io/Anvil.github.io/}{\textcolor{groundblue}{\faGlobe\enspace Project}}\qquad
\href{https://github.com/Knowin-29/Anvil/}{\textcolor{groundblue}{\faGithub\enspace Code}}\qquad
\href{https://huggingface.co/collections/Knowin-29/anvil-6ad0934f9fe868151f1ccb9a}{\textcolor{spatialteal}{\faDatabase\enspace Data}}\qquad
\href{https://huggingface.co/Knowin-29/Anvil-MOPD-9B}{\textcolor{actionorange}{\faCube\enspace Model}}}
\end{minipage}}

\makeatletter
\renewcommand{\@maketitle}{%
  \vbox{\hsize\textwidth\centering
    {\fontsize{19}{22}\selectfont\bfseries
      \raisebox{-5pt}[0pt][0pt]{\includegraphics[width=25pt]{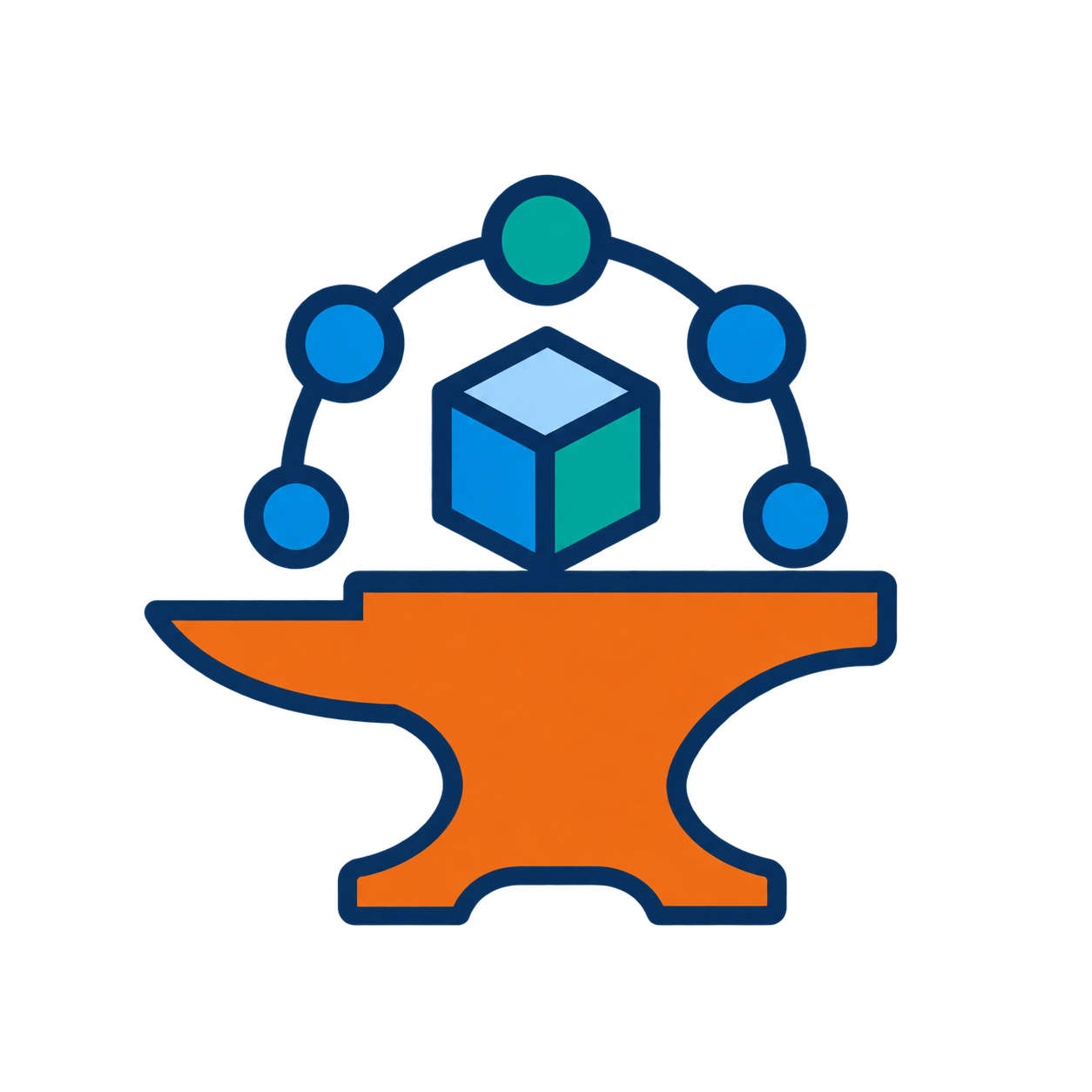}}\hspace{2pt}%
      \gradientAnvil: Scaling Embodied Spatial Intelligence\\
      through Knowledge Recirculation\par}%
    \vskip 11pt
    \@author\par
    \vskip 10pt
  }%
}
\makeatother

\renewenvironment{abstract}{%
  \begin{tcolorbox}[enhanced,colback=abstractwash,colframe=groundblue!22,
    boxrule=0.4pt,arc=3pt,boxsep=0pt,left=13pt,right=13pt,top=10pt,bottom=10pt,
    borderline west={2pt}{0pt}{titleblue},before skip=0pt,after skip=12pt]
  {\centering\large\scshape\color{titleblue}Abstract\par}\vspace{5pt}
  \noindent\ignorespaces
}{\par\end{tcolorbox}}

\begin{document}

\maketitle
\thispagestyle{anvilfirst}

\begin{abstract}
Embodied spatial intelligence requires more than visual recognition: models must localize entities, reason about their relations, and express this knowledge in action-relevant forms.
Yet joint training can conflate the knowledge needed at different stages because supervision is fragmented across incompatible geometric and semantic outputs.
We introduce \method, a unified framework for data generation, training, and benchmarking for embodied spatial intelligence.
\ding{182}\engine is a heterogeneous data flywheel that combines form-specific generation, cross-stage consistency verification, and failure-driven recollection across 11 Ground--Relation--Act forms.
\ding{183}\bench provides scene-aligned supervision for learning diverse embodied spatial capabilities and scene-linked assessments that distinguish stage competence from error propagation.
\ding{184}\model is a 9B student that consolidates heterogeneous spatial knowledge through distillation from specialist teachers.
Semantic-first-fault routing targets teacher supervision to the student's earliest decisive failure, while stage normalization controls the loss scale across native outputs of different lengths.
We construct 5.8M source-grounded records spanning 160K images,
reserving 580K spatial records for held-out evaluation
and using the remaining 5.22M records for training. On average across seven public benchmarks, \model outperforms Qwen3.5-9B and RynnBrain-8B by 9.29 and 11.03 percentage points, respectively, while reaching 80.67\% on VABench-Point. %
Controlled studies support targeted coverage, selective transfer, and feedback-guided repair as complementary framework components.
\end{abstract}

\section{Introduction}
\label{sec:intro}
Embodied spatial intelligence requires more than visual recognition: a model must localize entities, reason about relations, and identify possible interactions~\citep{yuan2024robopoint,roborefer2025,li2026rynnbrain11}. 
Recognizing a coffee machine, for example, does not determine where a cup can be placed beside it. Learning these capabilities requires supervision that connects visual observations to precise, action-relevant spatial judgments.
Web images, reconstructed scenes, and simulated environments offer complementary sources, but differ in their coverage and available geometry~\citep{song2024robospatial,spatialvlm2024,zhi2025sprite}.
\textit{Scaling embodied spatial intelligence therefore requires both broader supervision and a way to consolidate the knowledge expressed in different output forms.}

Recent data-driven approaches have advanced spatial grounding and reasoning through synthetic supervision, geometric reconstruction, and capability-diverse training~\citep{spatialvlm2024,roborefer2025,cai2026sensenovasi}.
Broader multimodal and embodied training also demonstrated the
value of integrating diverse visual experience and tasks~\citep{tong2024cambrian,eriq2025,galaxea2026g05,tencent2026hyvla,tan2026robobrain25,li2026rynnbrain11}.
These studies support scaling spatial supervision, although gains vary across capabilities and transfer settings~\citep{
cai2026sensenovasi,li2026rynnbrain11}.
Diagnostic benchmarks also revealed differences that aggregate
scores can obscure~\citep{liu2024embspatial,wu2025vsp,zhang2026spinbench}. Building on these advances, we investigate a data-centered question:
\begin{introquestions}
\emph{How can heterogeneous spatial expertise be turned into
reliable supervision and consolidated in a model that also
supports subsequent data verification?}
\end{introquestions}

\begin{figure}[!t]
\centering
\includegraphics[width=\linewidth]{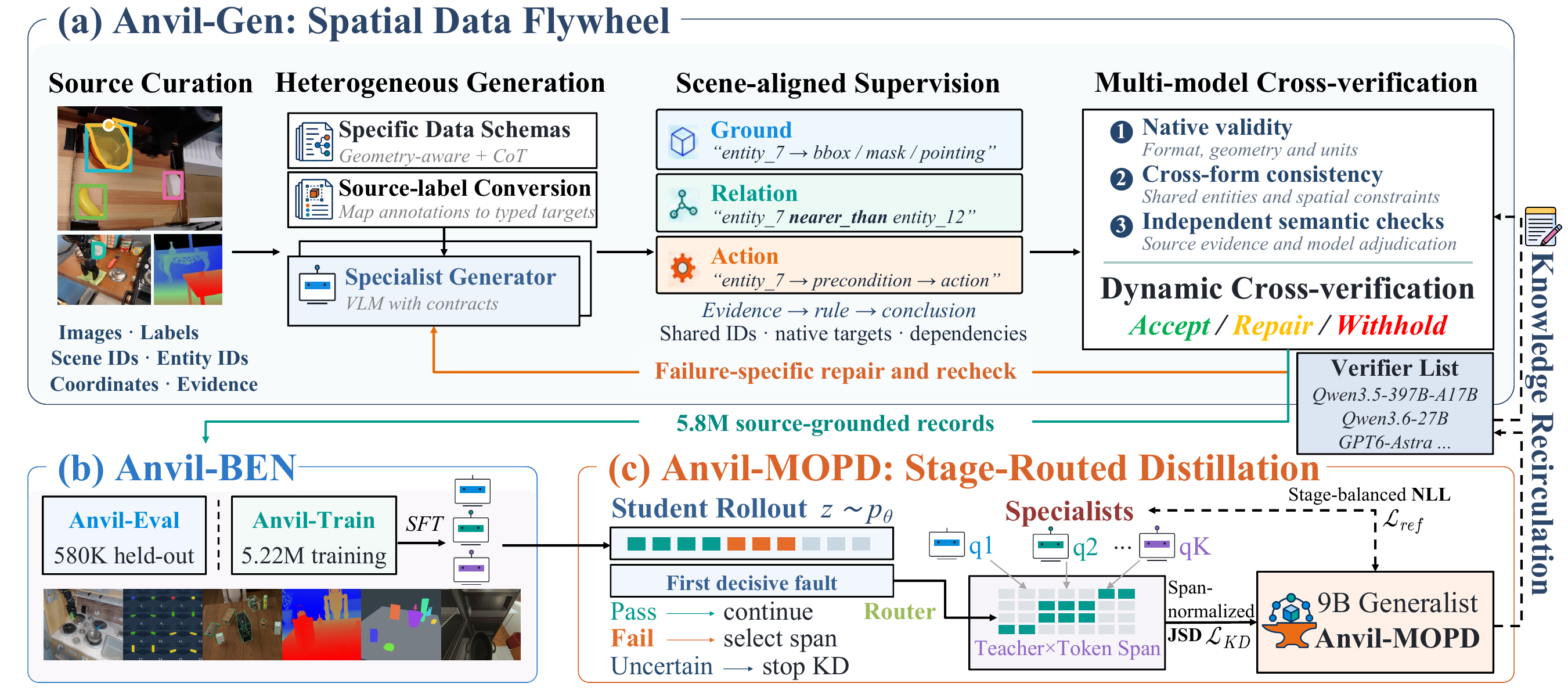}
\caption{\textbf{Overview of the Anvil framework.}
Anvil-Gen produces spatial supervision through generation,
cross-verification, and repair. Anvil-Ben curates verified records
into training and held-out evaluation sets. Anvil-MOPD consolidates
specialist knowledge into a generalist through teacher-and-span
routing on student rollouts. The generalist then assists subsequent
verification, returning learned knowledge to supervision production
while evidence-backed checks retain acceptance authority.
}
\label{fig:framework}
\label{fig:overview}
\label{fig:generation-cycle}
\label{fig:mtd-rich}
\end{figure}

We introduce \method, which connects a spatial data flywheel and model training through \textbf{knowledge recirculation} (Figure~\ref{fig:overview}): \engine operates as a heterogeneous data flywheel, where form-specific generation produces spatial supervision, cross-stage verification identifies inconsistencies, and failure feedback guides targeted data regeneration and collection.
A shared Ground--Relation--Act interface connects 11 native output forms through common entities and explicit dependencies, allowing the same capabilities to be specified across generation, learning, and evaluation.
\bench organizes the aligned supervision into \train for learning and \eval for held-out assessment through native-output and scene-linked tasks.
\model integrates knowledge from specialists trained on this supervision through multi-teacher distillation, building on on-policy
knowledge transfer~\citep{agarwal2024gkd,ma2026mopd,gao2026openmopd}.
Semantic-first-fault routing directs teacher supervision to the earliest decisive student error, while stage normalization
controls the loss scale across output lengths.
Validated records can refresh specialists, while the distilled generalist
returns verification signals to the data flywheel. Our contributions are threefold:

\textbf{(i) \engine: a scene-aligned, cross-verified data flywheel.}
Shared scene and entity references align source annotations and
specialist outputs, while geometry-aware CoT structures
schema-constrained generation. Cross-form adjudication turns
dependencies among outputs into explicit verification constraints,
and failure feedback guides targeted repair across sources and forms.

\textbf{(ii) \bench: a scene-aligned benchmark for spatial learning and evaluation.} Aligned scene images and 11 native annotation formats support training in \train and held-out assessment in \eval, covering spatial understanding, reasoning, and state-conditioned plan verification.
Native-output tasks measure individual capabilities; linked assessments and controlled upstream substitutions measure how errors propagate across dependent stages. %

\textbf{(iii) \model: specialist-to-generalist transfer with multi-teacher distillation.} Semantic-first-fault routing jointly selects a specialist and correction span, while stage normalization balances supervision across heterogeneous outputs. The resulting generalist can assist \engine under evidence-backed verification. Benchmark comparisons and controlled studies demonstrate the benefits of structured supervision and selective specialist transfer. %

\section{Related Work}
\label{sec:related}

\paragraph{Spatial intelligence.}
Spatial intelligence research spans relation recognition, geometric estimation, and reasoning about possible interactions. Early evaluations of relational understanding~\citep{liu2023vsr,kamath2023whatsup} have been complemented by image-based, multi-view, and video-based
benchmarks~\citep{tong2024cambrian,liu2024embspatial,fu2024blink,vsibench2024}, with further emphasis on transformations, planning, and spatial-functional reasoning~\citep{zhang2026spinbench,wu2025vsp,zhang2026embodied3dbench,zhang2026sfibench}. Training has likewise moved beyond image--text pairs toward supervision derived from geometry and robot experience~\citep{spatialvlm2024,cheng2024spatialrgpt,chen2025robo2vlm}, including precise, interaction-relevant grounding~\citep{yuan2024robopoint,roborefer2025}.
Procedural geometry and executable representations broaden the available targets~\citep{assouel2026pgt,liu2026primitives}, while program execution and trajectory consistency help check automatically generated annotations~\citep{zhi2025sprite,hu2024vpd,blank2026sparc}.
This expanding coverage raises a broader evaluation question: \textit{when does success on a particular spatial task indicate competence that transfers to other tasks?}

\paragraph{Embodied foundation models.}
Large-scale pretraining has connected visual--linguistic knowledge to embodied reasoning and robot control~\citep{driess2023palme,zitkovich2023rt2,kim2025openvla}. Autoregressive and diffusion policies provide alternative action interfaces, with generalist policies supporting adaptation across robot platforms~\citep{octo2024}.
In parallel, spatially focused models develop geometric and spatiotemporal understanding~\citep{cai2026sensenovasi,tan2026robobrain25,li2026rynnbrain11}, using structured scene representations, targeted supervision, or progressive curricula~\citep{ju2026momagraph,ray2025sat,li2026spatialladder}. Reasoning--action models increasingly connect these capabilities to action prediction~\citep{eriq2025,galaxea2026g05}.
\textit{The challenge is to turn diverse training signals into generalizable capabilities, rather than a collection of task-specific skills.}

\paragraph{Knowledge distillation.}
Distillation transfers teacher knowledge through output distributions or generated sequences~\citep{hinton2015distilling,kim2016sequencekd}; subsequent language-model methods address student behavior through reverse-KL and on-policy learning~\citep{gu2024minillm,agarwal2024gkd}.
Multi-teacher extensions introduce choices about both knowledge sources and supervision budgets~\citep{ma2026mopd,gao2026openmopd}.
Selection can operate over examples and tasks through rationale consolidation, answer verification, or privileged
context~\citep{jin2026knowledgepurification,he2026mtsdpo,ma2026softpromptopd}, or within trajectories through discrepancy-triggered intervention and failed-prefix recovery~\citep{koo2025switch,xu2026relayopd}.
Related approaches target informative tokens~\citep{liu2026vaopd,li2026crop},
cross-view representations~\citep{nguyen2026mvrd}, and intermediate reasoning steps~\citep{lightman2023prm}, while iterative bootstrapping reuses answer-checked model outputs for further training~\citep{zelikman2022star}.
Selective supervision thus raises a question of reliability: \textit{how can useful teacher knowledge be retained without also reinforcing its errors?}

To address these gaps, \method treats supervision reliability and knowledge transfer as coupled problems. Scene-aligned annotations support specialist-to-generalist distillation and held-out evaluation through knowledge recirculation; the resulting student can then assist further data verification. The knowledge recirculation mechanism links data refinement to model learning so that their effects on spatial competence can be studied together.

\section{Anvil: Learning through Knowledge Recirculation}
\label{sec:method}

\method connects spatial data production and model learning through
\textbf{knowledge recirculation} (Figure~\ref{fig:framework}). Shared entities and task dependencies align heterogeneous targets within a scene. These references support cross-form verification and localized repair in \engine, then identify the specialist and response span requiring correction during distillation. The resulting generalist assists subsequent verification, while evidence-backed checks retain acceptance authority and \bench reserves separate targets for evaluation.

\subsection{Scene-Aligned Supervision}
\label{sec:contracts}

Let $I\in\mathbb R^{H\times W\times3}$ be an RGB image, with height $H$ and width $W$, $x$ an instruction, and $E$ its entity inventory.
Contract $d$ specifies target space $\mathcal Y_d$, required fields, coordinates, and a semantic evaluator. Eleven contracts comprise four Ground forms (box, detection, mask, point), four Relation forms (depth order, metric comparison, direction, relation graph), and three Act forms (affordance, gripper configuration, action plan). A task graph $\mathcal G=(\mathcal S,\mathcal E)$ declares active stages $\mathcal S\subseteq\{G,S,A\}$ and the upstream fields consumed along
edges $\mathcal E$.

Stage $k$ has form $d_k$ and candidate target $\tilde y_k\in\mathcal Y_{d_k}$, giving $\tilde r=(I,x,E,\mathcal G,\{d_k,\tilde y_k\}_{k\in\mathcal S})$. Source annotations and specialist outputs are mapped to this interface using available coordinate and calibration metadata. Alignment preserves entity identities and supported units; it does not turn relative estimates into metric evidence. An atomic record contains one stage, whereas linked records expose the fields consumed downstream. Separating native representation from task dependency preserves the distinction between geometry and judgments derived from it, without requiring every scene to support all eleven forms (Appendix~\ref{app:schema}).

\begin{figure}[!t]
\centering
\includegraphics[width=\linewidth]{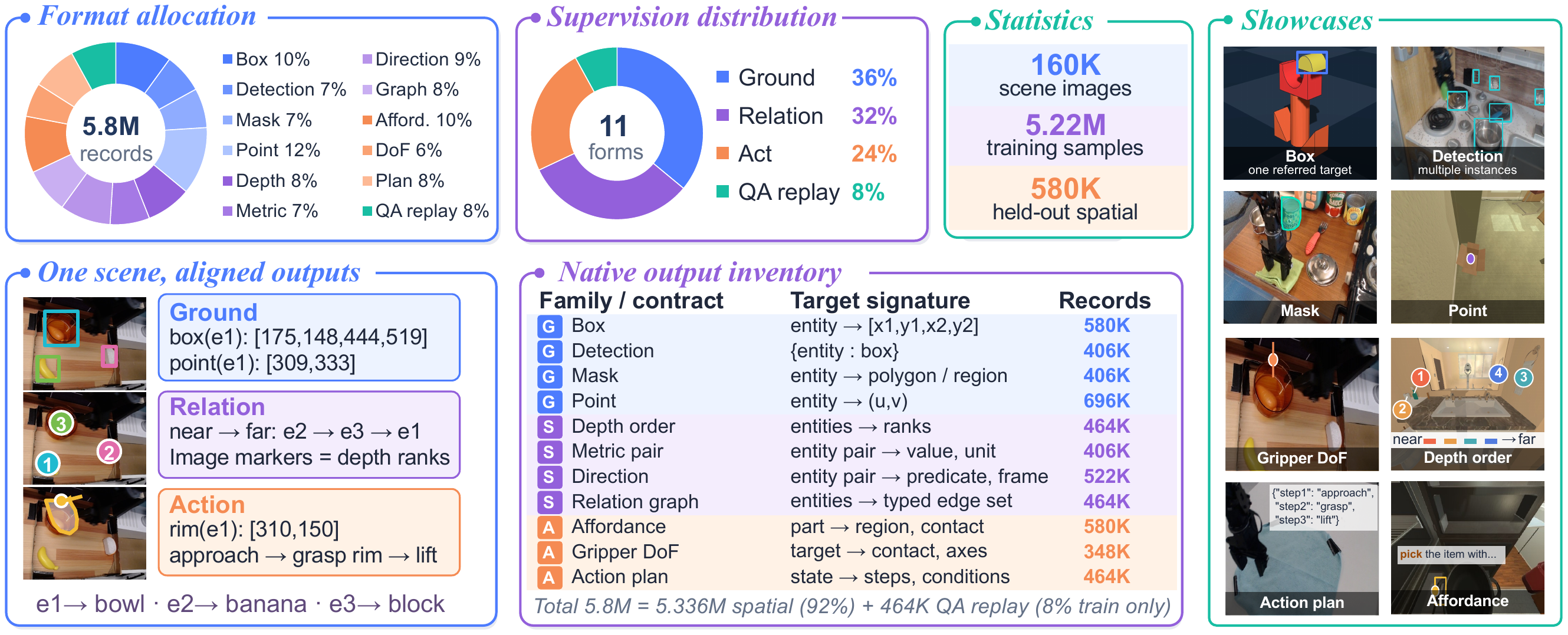}
\caption{\textbf{Scene-aligned supervision and training interfaces.}
The distributions summarize the full pool, including training-only
QA replay; the inventory lists eleven native spatial formats
and their target signatures. Shared entities connect the left example,
while the right gallery illustrates eight formats from
different scenes. Appendix Table~\ref{tab:scene-sources}
describes source provenance.}
\label{fig:anvil_data}
\label{fig:native-supervision}
\end{figure}

\subsection{Generation, Verification, and Feedback}
\label{sec:gate}
Native targets are obtained through source-annotation conversion or
schema-constrained specialist generation. Both paths retain scene and
entity identities for the shared admission interface.
For candidate $\tilde r$, binary $C_k(\tilde r)$ checks native fields and geometry. Let $\mathcal P(\tilde r)$ contain its declared cross-form constraints, including dependency edges and available geometric anchors.
Binary $C_c(\tilde r)$ checks constraint $c$: entity agreement, point--region containment, or mask--anchor consistency. Let $V(\tilde r)\in\{\mathrm{pass},\mathrm{uncertain},\mathrm{fail}\}$ be the semantic
decision supported by source annotations, calibrated measurements, action rules, or stronger-model adjudication. Admission requires
\begin{equation}
 \operatorname{Accept}(\tilde r)=
 \underbrace{\prod_{k\in\mathcal S}C_k(\tilde r)}_{\text{native validity}}
 \underbrace{\prod_{c\in\mathcal P(\tilde r)}C_c(\tilde r)}_{\text{cross-form agreement}}
 \underbrace{\mathds{1}[V(\tilde r)=\mathrm{pass}]}_{\text{semantic support}}.
 \label{eq:verification-interface}
\end{equation}
Here $\mathds{1}$ is the indicator and empty products equal one.
Mutually consistent outputs can still be wrong, motivating external evidence and explicit uncertainty rather than acceptance by agreement alone. Accepted targets become references $y_k^\star$ in record $r$.
Appendix~\ref{app:schema} describes the native interfaces and the
annotation review procedure.
Rejected candidates receive a failure category, not the reference answer. Repair preserves the scene, instruction, and entity IDs; aggregate failures guide source--task sampling. Accepted training records supply supervised targets and distillation references, whereas evaluation targets remain outside this feedback path. %

\paragraph{Constructing \bench.}
Figure~\ref{fig:native-supervision} summarizes the eleven native
spatial formats, their shared scene annotations, and the composition of the full data pool.
We stratified the 5.8M source-grounded records by source and native form, assigning scene groups to disjoint partitions:
580K spatial records form \eval, while \train contains the remaining 5.22M records, including all 464K QA replay records.

Records sharing a scene identity remain in one partition; where
scene identities are unavailable, grouping uses image identity.
The 160K scene images cover the entire pool.
Evaluation targets remain outside training and feedback.
Native-output tasks assess geometric and relational correctness against reference annotations, while linked tasks vary upstream context with the downstream target held fixed. \eval assesses both spatial intelligence and the reliability of model outputs as evidence for subsequent reasoning.

\subsection{Stage-Routed Multi-Teacher Distillation}
\label{sec:mtd}
\label{sec:teachers}

Student $p_\theta$, with parameters $\theta$, and specialists $q_G,q_S,q_A$ derive from native-supervised \basemodel checkpoints. Teachers share the student's vocabulary and remain frozen within each distillation run. Following on-policy distillation~\citep{agarwal2024gkd,ma2026mopd}, they score student-generated prefixes. Rather than routing by prompt domain alone, we use stage references to choose \textbf{both teacher and correction span}.
For a linked response $\hat z\sim p_\theta(\cdot\mid I,x)$, the evaluator assigns $a_k\in\{+1,0,-1\}$ for pass, uncertain, or fail; unreached stages receive zero.
In a declared topological order $\prec$, the first decisive fault 
\begin{equation}
 \rho(r,\hat z)=\min_{\prec}\bigl\{k\in\mathcal S:
 a_k=-1,\ a_j=+1\ \forall j\prec k\bigr\},
 \label{eq:first-fault}
\end{equation}
with $\rho=\varnothing$ when the set is empty; $j\prec k$ ranges over active stages. The selected teacher $q_\rho$ supervises token positions
$\widehat T_\rho$ of that stage in $\hat z$; uncertainty stops downstream attribution. Atomic records bypass this search and use their declared family. Appendix~\ref{app:router-details} defines structural failures and thresholds.

Let $\ell_{\rm ref}$ be the average target-token negative log-likelihood within each stage, then averaged equally across active stages. Let $\ell_{\rm KD}$ average the teacher--student generalized Jensen--Shannon divergence over the selected span, with nonnegative token weights summing to one.
Stage and span normalization prevent serialization length alone from setting supervision mass. Define $\widehat T_\varnothing=\varnothing$ and eligibility $e=\mathds{1}[|\widehat T_\rho|>0]$. For the accepted native-task training distribution $\mathcal D_{\rm train}$, \mbox{the objective is} %
\begin{equation}
 \min_\theta\ \mathcal L(\theta)=
 \lambda_{\rm ref}\mathbb E_{r\sim\mathcal D_{\rm train}}[\ell_{\rm ref}(r)]
 +\lambda_{\rm KD}\mathbb E_{r,\hat z}[\ell_{\rm KD}(r,\hat z)\mid e=1].
 \label{eq:objective}
\end{equation}
Here $\lambda_{\rm ref},\lambda_{\rm KD}\geq0$ set the loss scales; the second expectation uses training records and their student-sampled responses. Sampling, teachers, and routing are fixed during each update. With no eligible response, the transfer term is zero, but reference supervision remains active. Appendix~\ref{app:distillation-details} gives the exact losses and shows how the selector induces a normalized teacher--token allocation.

The two objectives serve complementary roles: reference learning
maintains supervision across all active stages, whereas routed
distillation concentrates specialist guidance on the selected failure.
This separation allows correction to be selective without discarding
the remaining task supervision. Normalization also keeps a long action
serialization from receiving greater loss weight solely because it
contains more tokens than a compact spatial answer.

\paragraph{Knowledge recirculation.}
Verified scene structure guides model correction, while learned competence returns to supervision production: %
\begin{equation}
(\mathcal D^{(n)},Q^{(n)})
\xrightarrow{\mathrm{distill}_{\rho,w}}
p_{\theta^{(n)}}
\xrightarrow{\text{assist verification}}
\widetilde{\mathcal D}^{(n+1)}
\xrightarrow{\text{accept \& merge}}
\mathcal D^{(n+1)}.
\label{eq:knowledge-cycle}
\end{equation}
Here $\mathcal D^{(n)}$ denotes the accepted training set and $Q^{(n)}$
the specialists frozen within round $n$; $\rho$ selects the correction
stage, and $w$ specifies normalized token weights within its response span.
The generalist supplies auxiliary signals for candidate verification and repair, producing $\widetilde{\mathcal D}^{(n+1)}$. 
Admission remains governed by $\operatorname{Accept}$ rather than student agreement alone; accepted records can support specialist updates before the next round.
Shared scene references thereby connect data admission, selective transfer, and knowledge return, while \eval remains outside training and feedback. Appendix~\ref{app:recirculation-details} formalizes this coupling.
The following experiments quantify the contributions of verified supervision, selective transfer, and feedback-driven repair within this recirculation framework.
\label{sec:inference}
At deployment, \model performs spatial inference without
specialist teachers or reference-aware routing.

\section{Experiments}
\label{sec:experiments}
We first evaluate \model against general-purpose and specialist models
on public spatial benchmarks (\textbf{RQ1: Performance}).
Controlled ablations and retention tests examine how supervision coverage and teacher allocation affect spatial learning and existing capabilities (\textbf{RQ2: Learning mechanisms}). 
We then measure the efficiency of source-based supervision construction and feedback-driven repair (\textbf{RQ3: Production efficiency}). Finally, task-structure and input controls test whether predictions depend on relevant scene evidence and task dependencies
(\textbf{RQ4: Evaluation validity}). Complete comparisons appear in Appendix~\ref{app:tradeoffs}.

\begin{table*}[!t]
\centering
\caption{Spatial and embodied-perception evaluation results
(\%, higher is better). %
}
\label{tab:spatial_benchmark}

\fontsize{9}{10.5}\selectfont
\renewcommand{\arraystretch}{1.08} %
\setlength{\tabcolsep}{2pt}        %
\setlength{\extrarowheight}{0pt}
\setlength{\aboverulesep}{3pt}
\setlength{\belowrulesep}{3pt}
\arrayrulecolor{black}

\providecommand{\needcheck}[1]{\textcolor{orange!85!black}{#1}}

\def\byear#1{\,{\fontsize{8}{9}\selectfont\citeyearpar{#1}}}

\def\bhdr#1#2{{%
  \renewcommand{\arraystretch}{1}%
  \begin{tabular}[c]{@{}c@{}}
    #1\\[-1pt]
    {\fontsize{7.5}{8.5}\selectfont\citeyearpar{#2}}
  \end{tabular}%
}}

\begin{tabularx}{\linewidth}{@{}>{\raggedright\arraybackslash}X*{9}{c}@{}}
\toprule

\diagbox[
  width=\linewidth,
  height=2.8\baselineskip,
  innerleftsep=1pt,
  innerrightsep=3pt,
  linewidth=0.3pt
]{\textbf{Model}}{\textbf{Benchmark}}
& \bhdr{CV-\\Bench}{tong2024cambrian}
& \bhdr{Emb-\\Spatial}{liu2024embspatial}
& \bhdr{Ref-\\Spatial}{roborefer2025}
& \bhdr{W2P}{yuan2024robopoint}
& \bhdr{BLINK\\Depth}{fu2024blink}
& \bhdr{ERQA}{deepmind2025erqa}
& \bhdr{VABench\\Point}{yuan2026fsd}
& \textbf{Avg.}
& \begin{tabular}[c]{@{}c@{}}Anvil\\-Eval\end{tabular}
\\
\midrule

\rowcolor{black!4}
\multicolumn{10}{@{}l}{\rule[-3pt]{0pt}{10pt}\hspace{3pt}\textit{Spatial and embodied models}}\\

Cosmos-Reason1-7B\byear{nvidia2025cosmosreason1} & 77.10 & 68.87 & 0.54 & 3.21 & 69.35 & 40.07 & 0.00 & 37.02 & 2.80 \\
Magma-8B\byear{yang2025magma} & 60.98$^\dagger$ & 64.59$^\dagger$ & 4.50$^\dagger$ & 10.89$^\dagger$ & 65.32$^\dagger$ & 25.73$^\dagger$ & 8.20 & 34.32 & 15.18 \\
RoboPoint-13B\byear{yuan2024robopoint} & 79.68 & 49.31$^\dagger$ & 22.87$^{\dagger\mathrm{L}}$ & 46.80$^\dagger$ & 61.29$^\dagger$ & 29.30 & 19.09 & 44.05 & 18.80 \\
RoboBrain2.0-7B\byear{cao2025robobrain2} & 85.75$^\dagger$ & 76.32$^\dagger$ & 32.50$^\dagger$ & 63.59$^\dagger$ & 84.68 & 30.31$^\dagger$ & 26.67$^\dagger$ & 57.12 & 45.94 \\
MiMo-Embodied-7B\byear{hao2025mimo} & 88.82$^\dagger$ & 76.24$^\dagger$ & 48.00$^\dagger$ & 63.60$^\dagger$ & 62.10 & 46.75$^\dagger$ & 46.93$^\dagger$ & 61.78 & 4.37 \\
RynnBrain-8B\byear{dang2026rynnbrain} & 87.87 & 79.78 & 59.20$^\dagger$ & 67.00 & 90.32 & 46.80$^\dagger$ & 11.67 & 63.23 & 55.98 \\
RoboBrain2.5-8B-NV\byear{tan2026robobrain25} & 94.58$^\dagger$ & 75.58$^\dagger$ & 60.50$^\dagger$ & 68.00 & 84.68 & 45.99 & 42.00 & 67.33 & 61.17 \\
RoboRefer-2B-RFT\byear{roborefer2025} & 92.20 & 75.34 & 52.00$^{\dagger\mathrm{L}}$ & 71.00$^\dagger$ & 84.32 & 46.52 & 33.34 & 64.96 & 39.25 \\
RoboRefer-8B-SFT\textsuperscript{D}\byear{roborefer2025} & 96.24$^{\dagger\mathrm{S}}$ & 72.53$^\dagger$ & 52.00$^{\dagger\mathrm{L}}$ & 70.00$^\dagger$ & 92.74$^\dagger$ & 52.33 & 27.12 & 66.14 & 52.21 \\

\addlinespace[4pt]
\rowcolor{black!4}
\multicolumn{10}{@{}l}{\rule[-3pt]{0pt}{10pt}\hspace{3pt}\textit{General-purpose models}}\\

GPT-4o\byear{hao2025mimo} & 78.63$^\dagger$ & 71.92$^\dagger$ & 8.78$^\dagger$ & 20.41$^\dagger$ & 78.23$^\dagger$ & 32.48$^\dagger$ & 13.67$^\dagger$ & 43.45 & 29.67 \\
Gemini-2.5-Pro\byear{hao2025mimo} & 84.59$^\dagger$ & 78.74$^\dagger$ & 38.16$^\dagger$ & 42.38$^\dagger$ & 87.90$^\dagger$ & 51.02$^\dagger$ & 27.92$^\dagger$ & 58.67 & 55.99 \\
Gemini-3-Pro-Preview\byear{tan2026robobrain25} & 92.00$^\dagger$ & 76.62$^\dagger$ & 65.50$^\dagger$ & 61.52 & 90.30 & 66.00 & 32.34 & 69.18 & 57.94 \\
Qwen2.5-VL-7B\byear{bai2025qwen25vl} & 75.40$^\dagger$ & 70.25$^\dagger$ & 38.00$^\dagger$ & 42.00$^\dagger$ & 79.00 & 38.80$^\dagger$ & 24.50$^\dagger$ & 52.56 & 3.21 \\
InternVL3.5-8B\byear{wang2025internvl35} & 81.46$^\dagger$ & 70.26$^\dagger$ & 16.80$^\dagger$ & 16.49$^\dagger$ & 79.84 & 41.00$^\dagger$ & 11.93$^\dagger$ & 45.40 & 48.46 \\
Qwen3-VL-8B-Instruct\byear{bai2025qwen3vl} & 86.69 & 78.87 & 53.50 & 34.32 & 87.90 & 44.25 & 46.33 & 61.69 & 55.16 \\
\basemodel\byear{qwen2026qwen35} & 87.15 & 75.58 & 61.67 & 47.97 & 90.32 & 49.83 & 42.33 & 64.98 & 56.87 \\

\midrule
\rowcolor{teal!6}
\textbf{\model\ (ours)} & \textbf{88.21} & \textbf{82.34} & \textbf{59.00} & \textbf{68.00} & \textbf{91.13} & \textbf{50.52} & \textbf{80.67} & \textbf{74.27} & \textbf{65.35} \\
\bottomrule
\end{tabularx}

\par\vspace{4pt}
\begin{minipage}{\linewidth}
\fontsize{7.5}{9}\selectfont
\raggedright

\textbf{Notes.}
Avg.\ is the mean of the seven public benchmarks, excluding our evaluation benchmark. $^\dagger$: published source-protocol results;
$^{\mathrm{D}}$: supports RGB-D input;
$^{\mathrm{L}}$: RefSpatial Location only;
$^{\mathrm{S}}$: CV-Bench subset aggregate excluding counting.

\end{minipage}
\end{table*}

\subsection{Experimental Setup}
\label{sec:setup}
Eight public benchmarks cover spatial understanding (CV-Bench~\citep{tong2024cambrian}, EmbSpatial~\citep{liu2024embspatial},
BLINK~\citep{fu2024blink}), referring and placement (RefSpatial~\citep{roborefer2025}, Where2Place~\citep{yuan2024robopoint},
RoboSpatial~\citep{song2024robospatial}), and embodied reasoning and pointing (ERQA~\citep{deepmind2025erqa}, VABench~\citep{yuan2026fsd}).
The baselines span general-purpose VLMs, spatial specialists, and
embodied models.
We train \model from \basemodel using native spatial supervision and multi-teacher distillation, and report the final checkpoint in the main table. Local comparisons share task-specific prompts, decoding settings, and scorers; published scores retain their source protocols.
Ablations control training updates and record exposures for coverage,
initialization, teachers, and maximum rollout budgets for routing,
and candidate pools and per-candidate call counts for repair. Coverage and routing comparisons each use two seeds; full settings
appear in Appendix~\ref{app:training}.
The main-table mean covers seven public tasks; RoboSpatial supplies
an additional coverage control. Earlier source-filtered checkpoints support the intervention analyses, while linked-source models support the routing study. These studies test individual
mechanisms rather than provide additional scores for the final model.
Published cross-protocol averages are descriptive; within-protocol
comparisons provide the controlled evidence.

\subsection{Spatial Intelligence Assessment and Coverage}
\label{sec:sota}

\paragraph{Observation 1: heterogeneous supervision transfers to spatial reasoning and pointing.} 
We first ask whether \model improves both spatial understanding and interaction-relevant prediction (\textbf{RQ1}).
Across the seven public benchmarks in Table~\ref{tab:spatial_benchmark}, the final checkpoint achieves a $74.27\%$ mean, exceeding \basemodel by $9.29$ points. On EmbSpatial, it outperforms Qwen3-VL-8B and RynnBrain-8B by $3.47$ and $2.56$ points, respectively, showing gains over both general-purpose and embodied baselines.
Excluding VABench-Point, the mean improvement over the base is $4.45$ points.
The gains span spatial judgments and location prediction, with a regression
on RefSpatial.
In a preliminary control study with a source-filtered model, we test
whether pointing gains rely on multiple predictions.
Scoring only the first point on the common $286$-item VABench subset
retains a $34.62$-point advantage over the base (Table~\ref{tab:expanded-local}), supporting a localization benefit
beyond additional guesses.

\begin{figure}[!t]
\centering
\includegraphics[width=\linewidth]{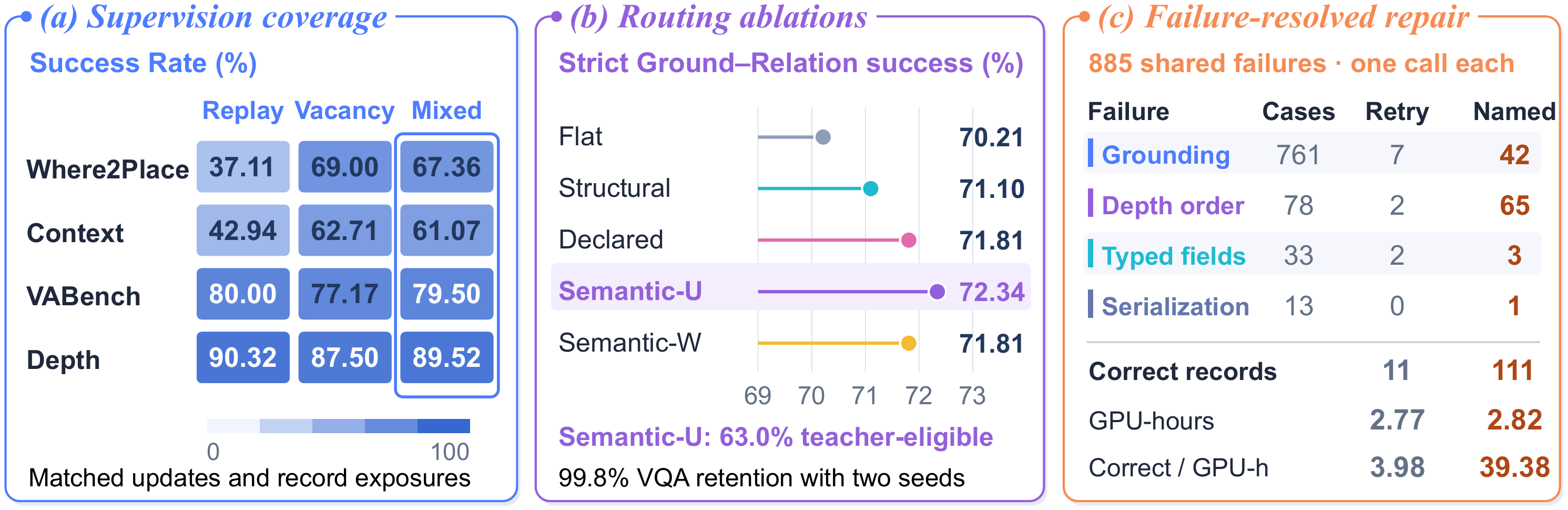}
\caption{\textbf{Controlled tests of coverage, routing, and repair.}
(a) Coverage matrix at matched updates and record exposures (Context: RoboSpatial; Depth: BLINK).
(b) Routing controls at matched maximum rollout budgets
(U: uniform; W: weighted).
(c) Correct repairs by failure type, using the same $885$ candidates
and one call each.}
\label{fig:mechanism-evidence}
\end{figure}

\subsection{Mechanism Ablations}
\label{sec:distillation-ablation}

\paragraph{Observation 2: targeted coverage improves adaptation, while replay mitigates forgetting.}
\label{sec:source-coverage}
We use preliminary controlled studies to isolate three decisions: which supervision to add, where to apply teacher correction, and how to repair rejected outputs (\textbf{RQ2}). %
Mixed training improves Where2Place over replay by $30.25$ points
across two seeds and also benefits RoboSpatial context
(Figure~\ref{fig:mechanism-evidence}a).
Fixed updates and exposures separate these gains from simply
training longer; increasing the decoding budget alone does not
recover placement performance.
Vacancy-only training favors placement,
while replay better preserves pointing and depth.
This trade-off motivates targeting \engine at missing capabilities
while retaining replay.
Mixed training does not improve every task, and unique images and realized tokens remain unmatched (Appendix~\ref{app:training}).

\paragraph{Observation 3: selective teacher correction improves transfer with limited retention loss.}
We test where teachers should intervene. With common initialization and a matched maximum rollout budget, semantic-uniform routing improves strict Ground--Relation success over flat distillation by $2.13$ points
(Figure~\ref{fig:mechanism-evidence}b). Semantic references expose incorrect but well-formed geometry that structural checks miss.
Teacher supervision reaches $63.00\%$ of sampled responses,
while reference learning covers all active stages, with $99.83\%$ general-VQA retention. The contrast supports selective correction over flat distillation.

\paragraph{Observation 4: failure-specific feedback improves repair yield at comparable cost.}
\label{sec:validity-evidence} We test whether failure-specific information improves repair. Using Qwen3.6-27B as the repair model, both arms receive the same $885$ rejected candidates and one call per candidate.
Named feedback recovers $10.09\times$ as many source-correct records
as generic retry at comparable GPU cost (Figure~\ref{fig:mechanism-evidence}c). It identifies the failed category without reference coordinates or ranks, preserving the task while localizing correction.
Recovery concentrates on depth ordering; grounding remains harder to repair.

Together, these controls support organizing supervision around
capability gaps, localizing teacher correction, and using explicit
failure information to recover usable training targets.

\FloatBarrier
\subsection{Qualitative Analysis and Production Efficiency}
\label{sec:production-efficiency}

\begin{figure}[!htbp]
\centering
\includegraphics[width=\linewidth]{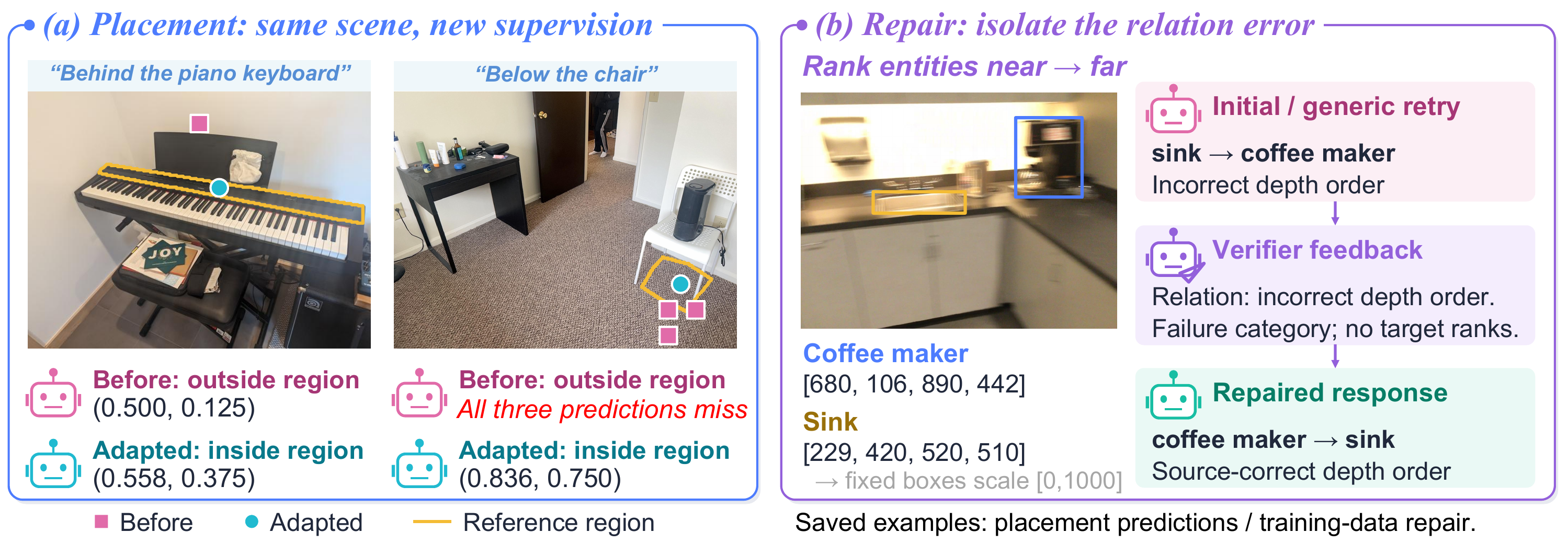}
\caption{\textbf{Qualitative examples of coverage adaptation and targeted repair.} 
(a) Saved placement predictions before and after adaptation (square: before; circle: adapted; yellow: reference region). 
(b) Verifier feedback guides a depth-order correction while object boxes remain unchanged. 
Selected successes illustrate the mechanisms, not their frequency.} %
\label{fig:placement-showcase}
\label{fig:feedback-efficiency}
\end{figure}

Figure~\ref{fig:placement-showcase} shows how the preceding mechanisms
change individual outputs. In panel (a), adaptation moves placement
predictions into the reference region; for the chair example, all three
original predictions miss that region. In panel (b), feedback corrects
the depth order of the coffee maker and sink while preserving their boxes.
The latter case illustrates how an error can be localized to Relation
while retaining the Ground output. These selected cases
explain the form of the correction; the ablations measure its aggregate
benefit, and Appendix~\ref{app:scale-evidence} includes a placement regression.

\paragraph{Observation 5: source evidence amortizes supervision construction.}
We next examine where supervision production incurs computational cost
(\textbf{RQ3}). From $9{,}608$ prepared RGB-D frames, eight CPU workers construct $2{,}463$ linked records in $325.25$ seconds without generative annotation. Registered depth and instance geometry provide the shared references needed to link Ground and Relation targets. A separate pass rechecks $598{,}891$ stored records in $13.26$ seconds, covering reading, parsing, hashing, and rule evaluation across four shards (Appendix~\ref{app:scale-evidence}). These measurements show that available source evidence supports both target construction and large-scale structural checking with CPU computation. They are component timings: source preparation,
image loading during rechecking, and semantic review remain outside the measured scope.

For candidates requiring model-based repair, the relevant efficiency measure is source-correct output per unit of computation. Normalizing the repair comparison by allocated GPU time gives $39.38$ correct records per GPU-hour for named feedback and $3.98$ for generic retry, a $9.90\times$ advantage. The images, questions, and candidate pool are fixed, so this gain reflects the value of localized error information under the same repair workload. Thus, we reuse source geometry, apply inexpensive structural checks, and direct model calls to unresolved semantic errors. 

The practical implication is to allocate computation according to
the evidence available for each record. Source-supported targets can
be constructed and checked directly; unresolved candidates receive
model-based adjudication or repair. This division makes heterogeneous
sources useful for different parts of the production process, rather
than requiring every annotation to pass through the same generative workflow.

\FloatBarrier
\subsection{Task Structure and Evidence Dependence}
\label{sec:dependency-evidence}
\begin{figure}[!htbp]
\centering
\includegraphics[width=\linewidth]{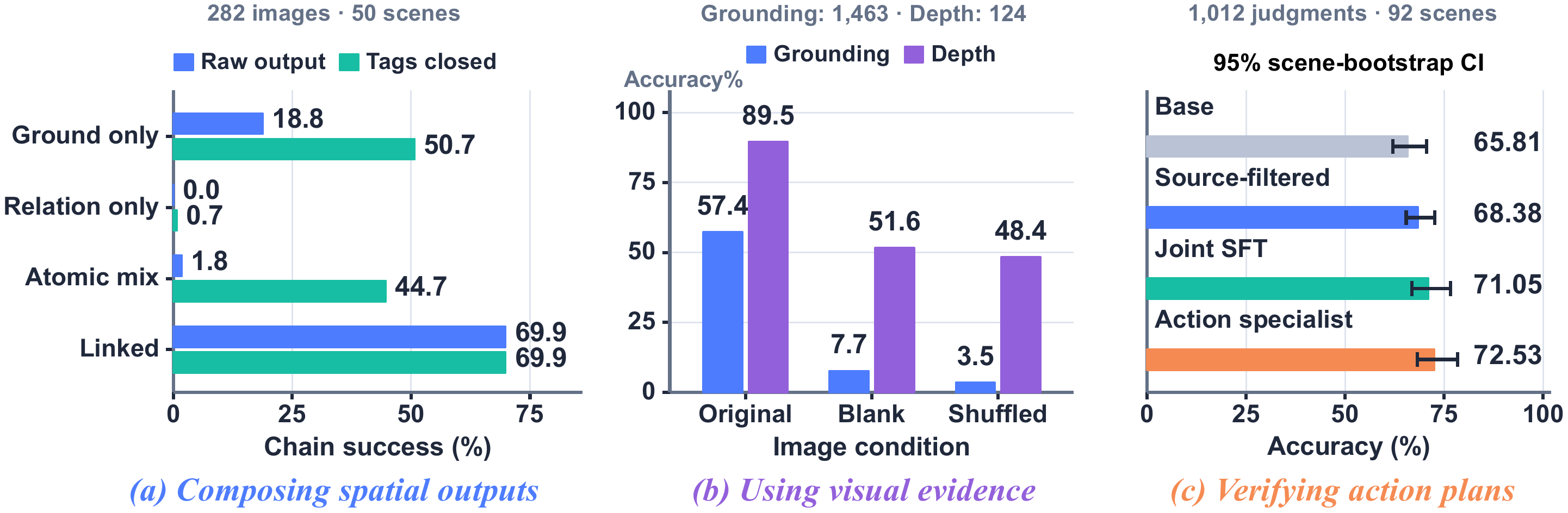}
\caption{\textbf{Task structure and evidence dependence.}
(a) Linked-task success before and after closing missing stage tags,
with JSON outputs unchanged.
(b) Grounding and depth under original, blank, and shuffled images.
(c) Plan-validity accuracy on $1{,}012$ constructed judgments with $95\%$
scene-bootstrap intervals. Panels use the study-specific checkpoints
described in Appendix~\ref{app:scale-evidence}.}
\label{fig:source-evidence}
\end{figure}

We ask whether linked-task gains survive format correction, whether predictions depend on visual and upstream evidence, and whether action judgments distinguish plans containing the same actions (\textbf{RQ4}). Figure~\ref{fig:source-evidence} addresses the question through composition, visual-input, and action-order controls.

\paragraph{Observation 6: scene-linked supervision substantially improves spatial composition.}
We first test whether independently learned Ground and Relation outputs compose under a shared interface. On $282$ images from $50$ held-out physical scenes, linked supervision reaches $69.86\%$ strict
Ground--Relation success, substantially exceeding the atomic mixture at matched updates (Figure~\ref{fig:source-evidence}a). To separate missing delimiters from errors in the predicted content, we close stage tags in the saved responses without changing their JSON outputs. This raises the atomic mixture to $44.68\%$ while leaving linked performance unchanged.
The remaining gap indicates that output syntax is only part of the
composition problem. A downstream stage must also receive the correct
entities and geometry from its predecessor. Scene-linked supervision
trains this dependency explicitly, making intermediate outputs useful
inputs to subsequent reasoning rather than merely separately correct answers.

\paragraph{Observation 7: spatial predictions depend on images and upstream geometry.}
We next examine dependence on the evidence supplied to these stages.
With questions and targets fixed, blank images reduce the
source-filtered model's accuracy on the 1,463-example grounding
panel from 57.35\% to 7.66\%. Cross-scene shuffling also degrades grounding, while BLINK-depth approaches chance (Figure~\ref{fig:source-evidence}b). Keeping images and depth targets fixed but shifting upstream boxes reduces the Relation specialist's accuracy by $10.64$ points (Table~\ref{tab:source-context}). In particular, errors in Ground can alter the context available to Relation even when the scene image remains fixed. These interventions establish sensitivity to both visual content and consumed geometry, motivating joint evaluation of stage accuracy and linked success.

\paragraph{Observation 8: action rules support state-conditioned plan verification.}
We extend the evaluation to action order under state constraints.
From $506$ ALFRED trajectories~\citep{shridhar2020alfred} across $92$
scenes, we construct $1{,}012$ balanced judgments, pairing each valid plan with an invalid permutation of the same actions. This fixes vocabulary and length, so action occurrence alone cannot distinguish the labels. Given the initial image, complete Boolean state, goal, and action rules, the Action specialist improves over the base by $6.72$ points (Figure~\ref{fig:source-evidence}c). Class-wise results remain uneven: accuracy is $89.53\%$ on invalid plans but only $55.53\%$ on valid ones (Table~\ref{tab:alfred-plan-verification}). The improvement mainly comes from recognizing more valid plans, although many feasible sequences are still rejected.

\paragraph{Summary and discussion.} 
The experiments suggest that spatial learning depends on how supervision
is organized, not only how much is available. Coverage controls favor task-relevant additions with replay; routing controls favor localized specialist correction; and repair controls show the value of actionable failure feedback. Source-based construction complements these mechanisms by reusing
available geometric evidence. Beyond aggregate benchmark gains, the composition, input, and action-order controls test whether predictions remain useful when later decisions depend on their content. These findings motivate scaling spatial supervision through capability coverage, selective correction, and explicit task dependencies.

\section{Conclusion}
\label{sec:conclusion}

We presented \method, which connects spatial data generation, evaluation,
and multi-teacher learning through knowledge recirculation. Shared scene references support cross-form verification, guide selective specialist transfer, and govern the acceptance of returned supervision.  The final model improves the seven-benchmark mean by 9.29 points over its base; controlled studies support targeted coverage, selective transfer, and feedback-driven repair. Together, these findings support jointly designing reliable supervision and knowledge consolidation for embodied spatial intelligence.

\label{main:end}

\FloatBarrier

\bibliography{references}
\bibliographystyle{anvil_author_year}

\FloatBarrier
\appendix
\raggedbottom
\definecolor{anvilAppBlue}{HTML}{0072B2}
\definecolor{anvilAppTeal}{HTML}{008564}
\definecolor{anvilAppOrange}{HTML}{C45B00}
\definecolor{anvilAppInk}{HTML}{233342}
\definecolor{anvilAppRed}{HTML}{A33A3A}
\colorlet{anvilAppHead}{anvilAppBlue!9}
\colorlet{anvilAppFocus}{anvilAppTeal!7}
\colorlet{anvilAppNeutral}{black!4}
\providecommand{\appbest}[1]{\textbf{#1}}
\providecommand{\appgain}[1]{\textcolor{anvilAppTeal}{\textbf{#1}}}
\newcommand{\appG}{\textcolor{anvilAppBlue}{\textbf{G}}}
\newcommand{\appS}{\textcolor{anvilAppTeal}{\textbf{S}}}
\newcommand{\appA}{\textcolor{anvilAppOrange}{\textbf{A}}}
\newcommand{\appYes}{\textcolor{anvilAppTeal}{\ding{51}}}
\newcommand{\appNo}{\textcolor{anvilAppRed}{\ding{55}}}
\newcommand{\appUncertain}{\textcolor{anvilAppOrange}{\textbf{?}}}
\newcommand{\appField}[1]{\textcolor{anvilAppInk}{\texttt{#1}}}
\newcommand{\appTableSetup}{%
  \centering\small\renewcommand{\arraystretch}{1.16}%
  \setlength{\tabcolsep}{4pt}}
\newcommand{\appGalleryPanel}[4]{%
  \includegraphics[width=.325\linewidth,trim=#1bp #2bp #3bp #4bp,clip]%
  {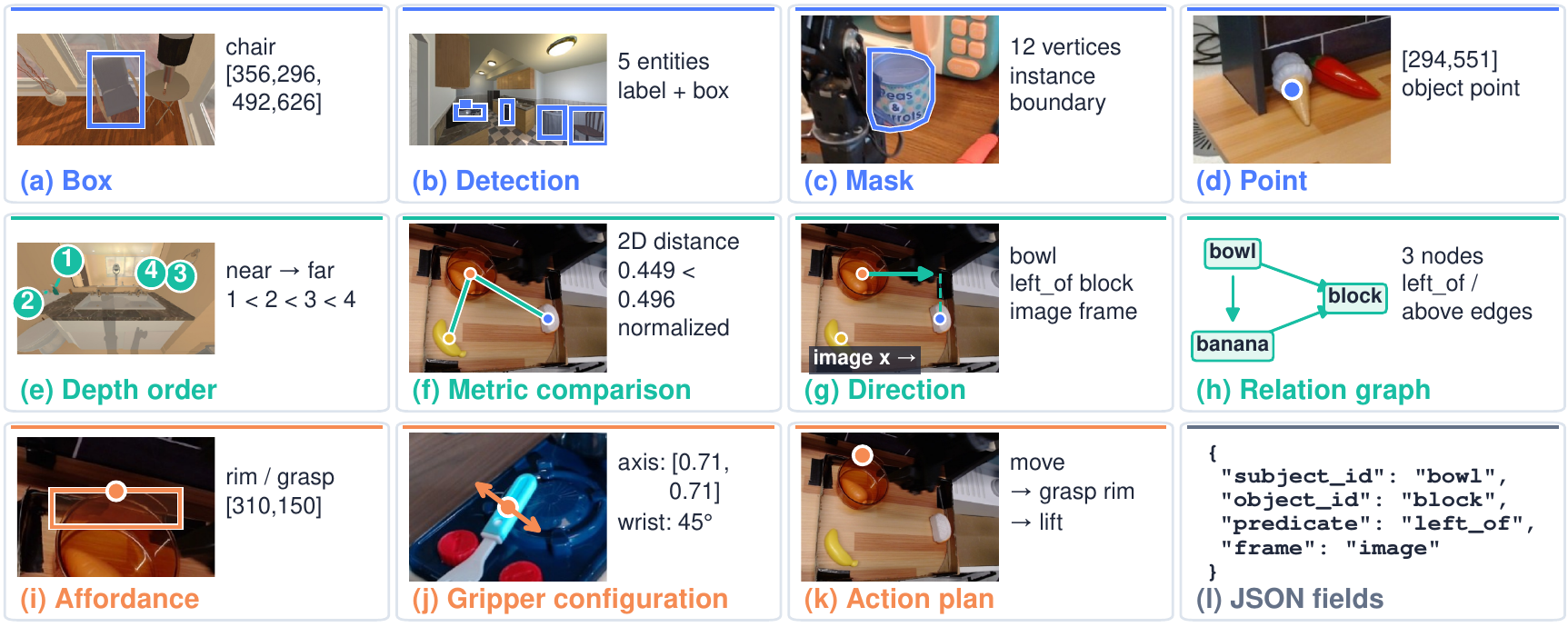}}
\newtcolorbox{anvilAppCard}[2]{enhanced,sharp corners,
  boxrule=0.4pt,colframe=#1!35,colback=#1!3,
  colbacktitle=#1!10,coltitle=anvilAppInk,fonttitle=\small\bfseries,
  title={#2},left=6pt,right=6pt,top=5pt,bottom=5pt,
  before skip=5pt,after skip=5pt}
\newtcolorbox{anvilAppNote}[1]{enhanced,sharp corners,
  frame hidden,borderline west={2pt}{0pt}{#1},colback=#1!5,
  left=7pt,right=7pt,top=5pt,bottom=5pt,
  before skip=6pt,after skip=6pt}
\renewcommand{\topfraction}{0.90}
\renewcommand{\bottomfraction}{0.80}
\renewcommand{\textfraction}{0.08}
\renewcommand{\floatpagefraction}{0.75}
\setcounter{topnumber}{3}
\setcounter{bottomnumber}{2}
\setcounter{totalnumber}{5}
\setlength{\emergencystretch}{1em}

\section{Spatial Data Flywheel and Native Supervision}
\label{app:schema}
This section details the supervision interfaces introduced in
Section~\ref{sec:contracts}, including native representations,
scene alignment, verification, and source provenance.
Routing, training, and evaluation protocols are provided in
Appendices~\ref{app:router-details}--\ref{app:scale-evidence}.

\subsection{Native interfaces and scene alignment}
\paragraph{Scene alignment.}
\label{app:scene-justification}
Each record retains the image, instruction, entity inventory, dependencies,
and native references from Section~\ref{sec:contracts}. Conversion
$\mathcal A_d(\xi;E,\kappa)\in\mathcal Y_d$ maps annotation $\xi$ to form $d$
using entity identities $E$ and calibration metadata $\kappa$.
Image coordinates $(u,v)$ become $(1000u/W,1000v/H)$.
Metric targets require supported units and calibration; relative estimates
remain relative. Shared identities connect available annotations without
requiring every scene to support all eleven forms or a universal sequence.
The reported $G\!\rightarrow\!S$ score measures composition between the grounding and spatial relationship.

\begin{table}[!htbp]
\appTableSetup
\caption{\textbf{Eleven native spatial interfaces.}
The symbols summarize output structure; colors and family letters identify
Ground (G), Relation (S), and Act (A). Each form retains its own validity
conditions, while shared entity references connect available forms.}
\label{tab:contracts}
\begin{tabularx}{\linewidth}{@{}p{.12\linewidth}p{.25\linewidth}X@{}}
\toprule
\rowcolor{anvilAppHead}
\textbf{Family} & \textbf{Form / structure} & \textbf{Native target and consistency requirement} \\
\midrule
\cellcolor{anvilAppBlue!7}\appG\ Ground & $\square$\ Box & Referred entity and bounded, ordered corners. \\
\cellcolor{anvilAppBlue!7} & $\{\square_i\}$\ Detection & Entity-indexed boxes with non-duplicate instances. \\
\cellcolor{anvilAppBlue!7} & $\{(x_i,y_i)\}$\ Mask & Instance or part region consistent with its anchor. \\
\cellcolor{anvilAppBlue!7} & $\bullet$\ Point & Object, contact, or free-space location in a declared frame. \\
\midrule
\cellcolor{anvilAppTeal!7}\appS\ Relation & $i\prec j$\ Depth order & Shared entity set and consistent near--far ordering. \\
\cellcolor{anvilAppTeal!7} & $d(i,j)$\ Metric comparison & Distances or comparisons with supported units. \\
\cellcolor{anvilAppTeal!7} & $i\!\to\!j$\ Direction & Relative direction with an explicit reference frame. \\
\cellcolor{anvilAppTeal!7} & $(E,\mathcal R)$\ Relation graph & Typed edges over the shared entity inventory. \\
\midrule
\cellcolor{anvilAppOrange!7}\appA\ Act & $(a,p)$\ Affordance & Action, functional part, and compatible contact region. \\
\cellcolor{anvilAppOrange!7} & $(p,\mathbf v)$\ Gripper & Contact, approach, and jaw geometry. \\
\cellcolor{anvilAppOrange!7} & $a_1\!\to\!a_2\!\to\!\cdots$\ Plan & Goal, ordered steps, entity references, and state constraints. \\
\bottomrule
\end{tabularx}
\end{table}

\paragraph{Native supervision examples}
\label{app:data-materials}\label{app:native-examples}

\begin{figure}[!htbp]
\centering
\includegraphics[width=\linewidth]{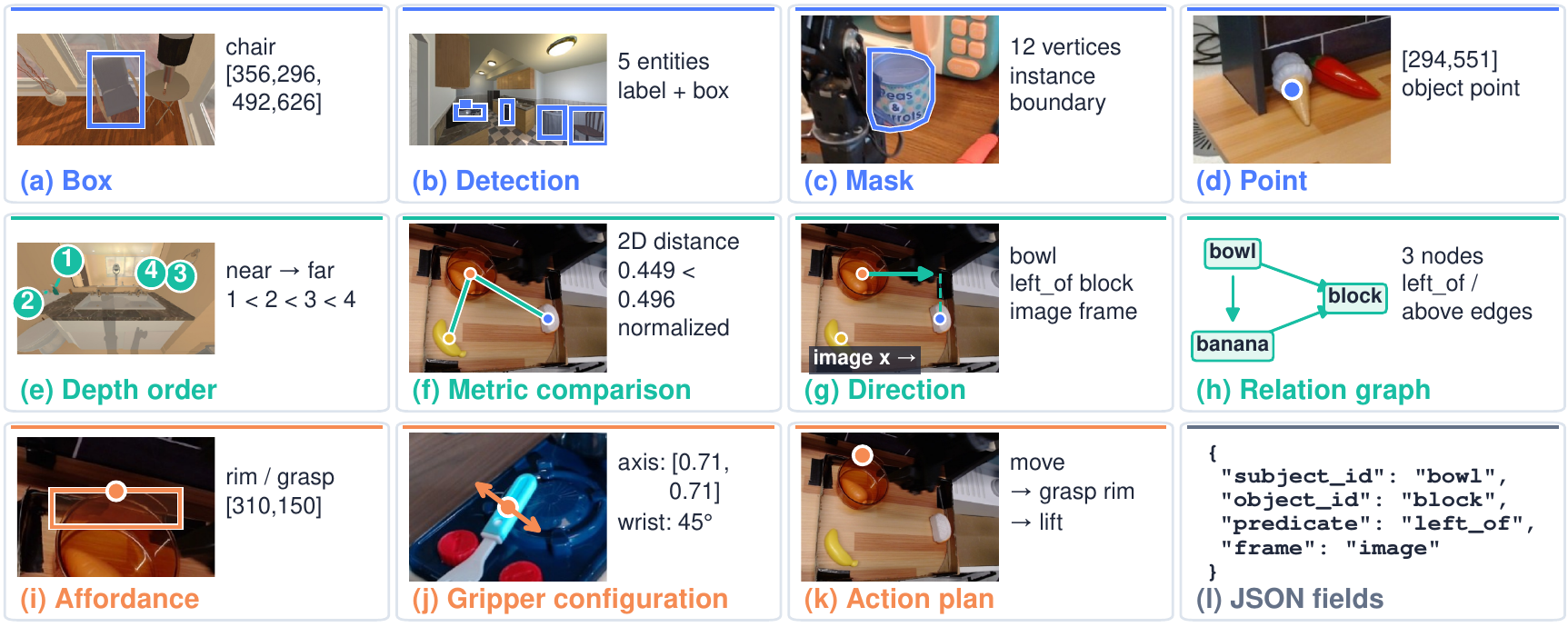}
\caption{\textbf{Examples of native spatial supervision.}
Panels (a--e) and (i--k) show stored annotations from an earlier
eight-form pilot. Panels (f--h) illustrate image-plane distance,
direction, and graph relations derived from saved entity boxes;
panel (l) shows fields rather than an additional task form.
Coordinates refer to the original full image on a $0$--$1000$ scale,
including when a panel displays a crop. The examples illustrate
recorded supervision and geometry-derived interfaces, rather than
model predictions or executed robot trajectories.}
\label{fig:material-gallery}
\end{figure}

Figure~\ref{fig:material-gallery} expands the compact examples in
Figure~\ref{fig:native-supervision}. The eight stored forms contain
object boxes, multiple detections, instance polygons, surface points,
depth ordering, contact regions, gripper configurations, and action
plans. These examples come from an earlier reviewed pilot and illustrate
the supervision interfaces; they are not predictions of \model.
Three additional panels illustrate relations derived from stored
image-plane geometry. Their distances use normalized image coordinates,
whereas physical distances require the calibration described in
Appendix~\ref{app:scene-justification}.

\paragraph{A shared contact across native forms.}
The orange-bowl example links a functional region to a plan through
the same contact point. The affordance record identifies the rim,
while the plan names that part in its grasp step and retains the
contact coordinates as its target. The following excerpts preserve
the stored field values; plan preconditions and failure modes are
omitted for brevity.

\begin{lstlisting}[basicstyle=\ttfamily\footnotesize,
  breaklines=true,columns=fullflexible,keepspaces=true,
  showstringspaces=false,frame=single]
Affordance:
{"label": "rim", "affordance": "grasp",
 "bbox_2d": [175,148,444,250],
 "contact_point_2d": [310,150]}

Action plan:
{"goal": "pick up the orange bowl",
 "steps": ["move to the orange bowl",
           "grasp the rim of the bowl",
           "lift the bowl upwards"],
 "target_point_2d": [310,150]}
\end{lstlisting}

Matching coordinates make the dependency inspectable. Semantic review
must additionally establish that the point lies on a visible contact
surface and that the stated action is supported by the available
evidence.

\paragraph{Generation constraints.}
\label{app:material-generation}

The archived generation templates make geometric requirements
explicit before requesting a native output. For example, the point
template asks for visible object material and warns that a box center
may lie in a hole or between chair legs. The mask template requires
a silhouette of the same grounded entity and permits an empty output
when a single polygon would include substantial background. The
depth template requires visible ordering evidence and permits
abstention for ambiguous pairs. These constraints provide concrete
examples of how the interfaces in Table~\ref{tab:contracts} enter
generation.

\begin{table}[!htbp]
\centering\small
\renewcommand{\arraystretch}{1.08}\setlength{\tabcolsep}{5pt}
\caption{\textbf{Representative generation constraints.}
Concise summaries of the archived templates; the requirements below
are not additional experimental measurements.}
\label{tab:material-prompt-constraints}
\begin{tabularx}{\linewidth}{@{}p{.19\linewidth}X@{}}
\toprule
\rowcolor{anvilAppHead}
\textbf{Form} & \textbf{Constraint supplied to the generator} \\
\midrule
Visible point & Preserve the grounded entity label; select visible material rather than a box center by default; return an empty array if no unambiguous point exists. \\
Mask & Segment the same grounded entity with 8--16 nested coordinate pairs; follow its silhouette rather than returning box corners; abstain when one polygon cannot represent it adequately. \\
Depth order & Preserve the entity set and labels; rank near to far using visible evidence; image height or apparent size alone is insufficient. \\
Contact region & Specify a visible functional sub-part and compatible action; place the contact on material rather than an empty container interior. \\
Gripper & Use the contact proposal, an image-plane jaw axis, an approach class, and a consistent wrist angle; abstain if secure contact is unsupported. \\
\bottomrule
\end{tabularx}
\end{table}

\paragraph{Review interface and decision rules.}
\label{app:review-workflow}

The annotation tools pair the source image and spatial overlays with
the candidate structured output. The review view exposes the relevant
entity context and records a support decision, an error category,
and an optional explanation. Unsupported decisions require an error
category. Stable sample identifiers and saved label snapshots bind
each decision to the image and candidate that were reviewed.

The archived review protocol specifies separate initial judgments
followed by adjudication of disagreements. It also distinguishes
operational inspection, which may display automatic verification
status, from the initial review view, which omits model confidence
and another reviewer's decision. Updating an image, target, or task
definition requires reviewing the affected candidate again rather
than transferring its previous decision.

\begin{table}[!htbp]
\centering\small
\renewcommand{\arraystretch}{1.08}\setlength{\tabcolsep}{5pt}
\caption{\textbf{Decision rules used to inspect spatial candidates.}
The checks concern the visible evidence and the declared task.
A valid serialization alone does not establish semantic support.}
\label{tab:material-review-rules}
\begin{tabularx}{\linewidth}{@{}p{.20\linewidth}XX@{}}
\toprule
\rowcolor{anvilAppHead}
\textbf{Target} & \textbf{Evidence to inspect} & \textbf{Reason to withhold or reject} \\
\midrule
Box / detection & Identity and visible extent of the requested entity. & Wrong instance, omitted material parts, or duplicated detections. \\
Mask / point & Visible object material and its boundary. & Background inside a polygon; a point in a hole or behind an occluder. \\
Depth order & Occlusion, contact layering, or a supported perspective cue. & A reversed pair or insufficient evidence to order it. \\
Affordance / gripper & Named contact part, contact point, and jaw geometry. & A grasp in empty container space or on an unsupported contact surface. \\
Action plan & Referenced entities, stated preconditions, and ordered steps. & Inapplicable actions or a target inconsistent with the named contact. \\
\bottomrule
\end{tabularx}
\end{table}

Review decisions and automatic admission results have different
denominators. Stored pilot judgments describe the candidates in that
pilot; a held-out audit must use the sample identities and protocol of
that held-out partition.

\paragraph{Concrete rejection cases.}
\label{app:material-failures}

The stored judgments include cases in which individually well-formed
outputs fail a physical or semantic check. In one pot scene, an
affordance names the rim while its contact point lies in the dark
opening. The gripper record and action plan reuse that location.
Agreement across the three outputs therefore preserves the same
unsupported contact. In another scene, a proposed grasp targets the
center of a sticker adhered to a board, without a pinchable edge.
A grounding example identifies a lamp but omits its visible base.
These cases motivate inspecting the referenced entity and contact
surface in addition to checking the output fields.

\paragraph{Source provenance and benchmark scope.}
\label{app:scene-sources}\label{app:benchmark-details}\label{sec:benchmark}
\engine constructs 5.8M source-grounded records over 160K images: 5.336M spatial records and 464K QA replay records. We reserve 580K spatial records for held-out evaluation and use 5.22M records for training, including all QA replay. Evaluation targets remain outside training and feedback. Table~\ref{tab:scene-sources} summarizes the source families. The controlled studies use task-specific evaluation panels: 1,463 box/point examples, 1,024 masks, 282 linked images from 50 scenes, and 1,012 plan judgments from 92 scenes. Each reported score uses its panel's denominator.
Historical human review and the new source-grounded evaluation partition are distinct. The 1,463-example grounding panel is a model-evaluation panel, not a human audit of the 580K partition; its size does not establish a corpus-wide annotation-accuracy estimate.

\begin{table}[!htbp]
\centering\small
\renewcommand{\arraystretch}{1.08}\setlength{\tabcolsep}{5pt}
\caption{\textbf{Source families and supervision roles.} MIMIC-CGD denotes the CGD subset of MIMIC-IT; VSI-590K is the instruction-tuning corpus released with Cambrian-S.}
\label{tab:scene-sources}

\begin{tabularx}{\linewidth}{@{}>{\raggedright\arraybackslash}p{.17\linewidth}>{\raggedright\arraybackslash}p{.23\linewidth}>{\raggedright\arraybackslash}X@{}}
\toprule
\rowcolor{anvilAppHead}
\textbf{Source family} & \textbf{Supervision role} & \textbf{Illustrative provenance} \\
\midrule
\textbf{Robot video} & Grounding, interaction & BridgeData V2~\citep{bridgedatav2} \\
\textbf{Embodied scenes} & Spatial targets & RefSpatial~\citep{roborefer2025} \\
\textbf{Simulation} & Objects, actions, physical relations & Embodied-Reasoner~\citep{embodiedreasoner2025}; PhyBlock~\citep{phyblock2025}; MineAnyBuild~\citep{mineanybuild2025} \\
\textbf{Spatial imagery} & Placement, spatial QA, relations & FSD~\citep{yuan2026fsd}; SAT~\citep{ray2025sat}; VSI-590K~\citep{cambrians2025}; SpatialSense~\citep{spatialsense2019} \\
\textbf{Object states} & State-conditioned grounding and actions & ChangeIt frames~\citep{changeit2022} \\
\textbf{Pointing} & Points and caption QA & PixMo~\citep{molmopixmo2025} \\
\textbf{QA replay} & Paired-image and compositional QA & MIMIC-CGD~\citep{mimicit2023}; NLVR2~\citep{nlvr22019}; CLEVR~\citep{clevr2017}; TallyQA~\citep{tallyqa2019}; VSR~\citep{liu2023vsr} \\
 & General visual QA & Localized Narratives~\citep{localizednarratives2020}; VQAv2~\citep{vqav22017}; COCO-QA~\citep{cocoqa2015}; A-OKVQA~\citep{aokvqa2022}; OK-VQA~\citep{okvqa2019}; Visual7W~\citep{visual7w2016} \\
\bottomrule
\end{tabularx}

\end{table}

\paragraph{Provenance, grouping, and record accounting}
\label{app:material-provenance}

Source conversion, model-generated candidates, and review decisions
carry distinct provenance. A converted reference is supported by its
source annotation or measurement; a generated candidate is associated
with the generator and template version; a review decision is tied
to a specific saved candidate. Geometry-derived examples additionally
declare their coordinate frame and units. This distinction prevents
an image-plane relation from being interpreted as a calibrated physical
measurement.

Image identity, scene grouping, and target identity address different
forms of duplication. Multiple native targets from the same scene
remain associated with their shared group when a scene identity is
available; otherwise, the grouping unit is the image, as specified in
Section~\ref{sec:gate}. A count of generated records can therefore
exceed the number of images and does not count independent scenes.
The figures illustrating formats are selected examples rather than
a representative sample from which corpus proportions are estimated.

\section{Routing, Distillation, and Knowledge Return}
\label{app:router-details}
\paragraph{Stage decisions.}
A reached stage with malformed output or a violated native constraint has $a_k=-1$. Otherwise, its native score $s_k$ fails below $\tau_k-\delta_k$, passes at or above $\tau_k+\delta_k$, and is uncertain between these bounds.
The source-linked study uses $(\tau_G,\delta_G)=(0.75,0.125)$ and
$(\tau_S,\delta_S)=(0.5,0.25)$, giving boundaries $0.625/0.875$ and $0.25/0.75$. Equality at the lower boundary is uncertain; equality at the upper boundary passes. Unreached stages are uncertain. The scan advances through passed stages. A definite failure selects that stage; uncertainty stops attribution without selecting a later stage.
Malformed outputs use the nonempty suffix after the last valid boundary as the correction span. 
Reference learning remains active for every record.
For linked records, distillation is omitted when the scan selects
no stage or the selected correction span is empty.
Atomic records use their declared family and sampled response span.

For linked tasks, a downstream stage consumes upstream fields,
so an upstream error can invalidate its input context.
First-fault routing therefore selects a downstream specialist only
after preceding stages pass their checks. Once an upstream fault
is selected, changing only later predictions cannot redirect correction.

\subsection{Normalized teacher--token allocation}
\label{app:distillation-details}
Let $\widehat T_k$ be the sampled positions of stage $k$, and let $w_{k,t}$
be nonnegative weights summing to one on a nonempty span and zero outside it.
The selector jointly allocates a specialist and correction tokens:
\begin{equation}
 \Omega_{k,t}=\mathds{1}[\rho(r,\hat z)=k]w_{k,t},\qquad
 \ell_{\rm KD}=\sum_{k\in\mathcal S}\sum_{t\in\widehat T_k}
 \Omega_{k,t}D_{\beta,t}^{(k)}.
\end{equation}
Here $D_{\beta,t}^{(k)}$ compares student and teacher $q_k$ on the same
student prefix. The matrix $\Omega\in\mathbb R_{\geq0}^{|\mathcal S|\times|\hat z|}$
has at most one nonzero row and total mass $e\in\{0,1\}$. This is the
implemented allocation, not an additional trainable module.

Student and teacher distributions share the union $U_t$ of teacher top-$K$
vocabulary entries and structural tokens realized in the selected span.
A residual category collects omitted probability. The resulting vectors
$\tilde p_{\theta,t},\tilde q_{k,t}\in\mathbb R^{|U_t|+1}$ retain that mass.
With stop-gradient $\operatorname{sg}$ and
$m_t=\beta\operatorname{sg}[\tilde q_{k,t}]+(1-\beta)\tilde p_{\theta,t}$,
\begin{equation}
 D_{\beta,t}^{(k)}=\beta\,\mathrm{KL}(\operatorname{sg}[\tilde q_{k,t}]\Vert m_t)
 +(1-\beta)\,\mathrm{KL}(\tilde p_{\theta,t}\Vert m_t),
 \qquad K=128,\quad\beta=0.5.
\end{equation}
Let $\mathcal K_k$ contain structural positions (fields, coordinates,
relation labels, action keywords, and delimiters). For $t\in\widehat T_k$,
\begin{equation}
 w_{k,t}=\frac{1+\alpha\mathds{1}[t\in\mathcal K_k]}
 {\sum_{j\in\widehat T_k}(1+\alpha\mathds{1}[j\in\mathcal K_k])}.
\end{equation}
Uniform and weighted variants use $\alpha=0$ and $\alpha=2$.
If $z^\star$ serializes references and $T_k^\star$ is its nonempty stage span,
\begin{equation}
 \ell_{\rm ref}(r)=-\frac{1}{|\mathcal S|}\sum_{k\in\mathcal S}
 \frac{1}{|T_k^\star|}\sum_{t\in T_k^\star}
 \log p_\theta(z_t^\star\mid I,x,z_{<t}^\star).
\end{equation}
The minibatch teacher loss is averaged over eligible examples
and is zero when none qualify.
Sampling, teachers, support selection, and routes are fixed within an update;
no policy gradient passes through the sampled response.
The residual probability $1-\sum_{v\in U_t}p(v)$ retains omitted mass.
Teacher positions count scored logits, including distributed padding,
separately from eligible responses, useful tokens, and wall time.

\subsection{Why normalize the correction span?}
\label{app:allocation-justification}
For eligible responses, $\Omega$ has unit mass on the selected span.
Reference learning assigns equal aggregate weight to active stages.
Both normalizations control loss coefficients independently of
serialization length.

The transfer term also has a length-independent bound. For
$0<\beta<1$, define
$h(\beta)=-\beta\log\beta-(1-\beta)\log(1-\beta)$.
Since $m_t(v)\geq\beta\tilde q_{k,t}(v)$ and
$m_t(v)\geq(1-\beta)\tilde p_{\theta,t}(v)$,
\begin{equation}
 0\leq D_{\beta,t}^{(k)}\leq h(\beta),\qquad
 0\leq\ell_{\rm KD}\leq e\,h(\beta).
 \label{eq:appendix-kd-bound}
\end{equation}
The upper bound follows by substituting the two mixture inequalities into
the KL terms and using $\sum_{k,t}\Omega_{k,t}=e$; nonnegativity follows
from KL nonnegativity. At $\beta=0.5$, it becomes $e\log 2$ in natural-log
units. This bounds loss magnitude, not gradient norms or downstream error.

\subsection{Round-level coupling}
\label{app:recirculation-details}
Let $\mathcal U_{\rm tr}$ denote training sources and $\mathcal V^{(n)}$
the verification ensemble used in round $n$. The ensemble includes the
strong-model verification stage and the generalist models used for
cross-checking. Let $\mathcal C^{(n+1)}$ be the candidates generated or
repaired with auxiliary student signals:
\begin{align}
 \mathcal C^{(n+1)}
 &=\mathsf{GenRepair}(\mathcal U_{\rm tr};p_{\theta^{(n)}}),\nonumber\\
 \Delta\mathcal D^{(n+1)}
 &=\{\tilde r\in\mathcal C^{(n+1)}:
       \operatorname{Accept}(\tilde r;\mathcal V^{(n)})=1\},\nonumber\\
 \mathcal D^{(n+1)}
 &=\mathcal D^{(n)}\cup\Delta\mathcal D^{(n+1)},\nonumber\\
 \theta^{(n+1)}
 &=\mathsf{Train}_{\Omega}
   (\theta^{(n)};\mathcal D^{(n+1)},Q^{(n)}).
\end{align}
Here $\operatorname{Accept}(\cdot;\mathcal V^{(n)})$ applies
Eq.~\ref{eq:verification-interface} using the round's verification
ensemble, and the union counts unique records. 
Student signals assist candidate generation or repair, while
admission remains governed by Eq.~\ref{eq:verification-interface}.
The symbols distinguish the verification role of $\mathcal V^{(n)}$ from the distillation role of $Q^{(n)}$. Distillation specialists remain frozen within each training run. Accepted records may support specialist refresh between runs. Evaluation targets remain outside this process; the experiments measure its component effects.

\section{Training and Evaluation Protocols}
\label{app:training}\label{app:configuration-map}

\paragraph{Checkpoints and study roles.}
The main table reports the final \model on seven public tasks;
RoboSpatial provides an additional coverage control. The coverage study
continues an earlier source-filtered checkpoint with replay, vacancy, or
mixed supervision, while routing continues a common linked-SFT checkpoint.
Composition and data-quality comparisons start from the base model.
The image interventions compare the base and the earlier source-filtered
\method checkpoint; box interventions compare the Relation specialist and
joint SFT. Plan verification additionally evaluates the Action specialist.
Throughout the appendix, \method\textsuperscript{$\dagger$} denotes the
\emph{earlier source-filtered checkpoint, not the final main-table model}.
The Qwen3.6-27B repair model has a separate production role: both repair
arms process the same 885 failures with one call per candidate.

\paragraph{Evaluation and uncertainty.}
Local comparisons fix prompts, decoding, and scorers; published values
retain the modalities and subsets marked in the main table.
Strict linked success requires valid serialization and correct dependent
outputs; plan validity uses exact VALID/INVALID classification.
Malformed or incomplete responses remain in the denominator.
Paired scene-bootstrap intervals use 5,000 resamples, keeping all
records from a sampled physical scene together and using identical
scene draws for the compared models. For routing, per-image outcomes
are first averaged across the two recorded seeds. The resulting
intervals condition on those seeds and do not estimate variability
over unobserved initializations.

\paragraph{Record-level aggregation on \eval.}
The \eval column in Table~\ref{tab:spatial_benchmark}
reports the record-weighted mean over the held-out evaluation
partition, excluding training-only QA replay.
Let $\mathcal D_{\mathrm{eval}}=\{r_i\}_{i=1}^{N_{\mathrm{eval}}}$
contain $N_{\mathrm{eval}}=580{,}000$ records.
For record $r_i$, let $d_i$ denote its native form,
$\hat y_i$ the model prediction, and $y_i^\star$ the reference.
Its score
$s_i=g_{d_i}(\hat y_i,y_i^\star)\in[0,1]$
is computed by the corresponding native-task evaluator.
Malformed, missing, or incomplete predictions receive zero
and remain in the denominator. The aggregate score is
\begin{equation}
S_{\mathrm{eval}}
=
\frac{100}{N_{\mathrm{eval}}}
\sum_{i=1}^{N_{\mathrm{eval}}} s_i .
\label{eq:eval-record-average}
\end{equation}
Each record contributes equally, regardless of its output length.
Consequently, forms contribute in proportion to their evaluated
record counts; the score is not an unweighted average across
the eleven forms or across scene images.
The reported $65.35$ for \model is this record-level aggregate,
expressed on a 0--100 scale.

\paragraph{Controlled-panel metrics.}
The mechanism studies use separate task-specific panels,
whose scores are reported with their own denominators rather
than substituted for the aggregate \eval score.
For a linked evaluation of $N$ images, strict success
can be written as
\begin{equation}
 S_{G\to S}=\frac{100}{N}\sum_{i=1}^{N}
 \mathds{1}[\text{valid serialization}_i]
 \mathds{1}[\text{Ground correct}_i]
 \mathds{1}[\text{Relation correct}_i].
 \label{eq:appendix-linked-score}
\end{equation}
Missing or incorrect stages contribute zero. Tag-closure analysis changes
only recognition of the saved stage boundaries; it does not replace predicted
geometry or relation labels. For pointing, any-point scoring succeeds when
at least one returned point satisfies the task criterion, whereas first-point
scoring evaluates only the first point in the same saved response. Plan
accuracy uses all $1{,}012$ judgments, with $506$ examples in each class.

\paragraph{Training configurations.}
Table~\ref{tab:appendix-training-settings} collects the optimization settings
for the controlled studies. These local continuation and fresh-base runs are distinct from
the final model reported in the main table.

\begin{table}[!htbp]
\centering\small
\renewcommand{\arraystretch}{1.08}\setlength{\tabcolsep}{4pt}
\caption{\textbf{Optimization settings for the controlled training studies.}
Batch denotes the configured effective batch size in records.
Coverage exposure counts and realized teacher positions are reported separately.}
\label{tab:appendix-training-settings}\label{tab:checkpoint-study-map}
\begin{tabular}{@{}lrrrr@{}}
\toprule
\rowcolor{anvilAppHead}
\textbf{Study} & \textbf{Updates} & \textbf{Batch} & \textbf{Learning rate} & \textbf{Seeds} \\
\midrule
Coverage continuation & 303 & 24 & $2\times10^{-5}$ & 43, 44 \\
Atomic / linked source SFT & 512 & 4 & $10^{-4}$ & 17 \\
Routing continuation & 256 & 4 & $5\times10^{-6}$ & 17, 23 \\
Matched-size / nested-pool SFT & 256 & 4 & $10^{-4}$ & 17 \\
\bottomrule
\end{tabular}
\end{table}

\paragraph{Coverage controls.}
Coverage arms use LoRA rank/scaling 16/32 and share a 360-record development
split. Single-source arms repeat 3,656 replay or 3,608 vacancy records to
match the reported 7,264 mixed-arm row exposures. Terminal checkpoints are
fixed before evaluation. Unique images and realized tokens are not matched.
A representative mixed run takes 2.37 L20 GPU-hours. Repetition changes the
source composition at the stated exposure budget.

\paragraph{Linked SFT and routing.}
Ground-only, Relation-only, atomic-mixture, and linked SFT use LoRA
rank/scaling 64/128, dropout 0.05, a frozen vision encoder, a maximum sequence
length of 3,072, and at most 1,536 image tokens. Cosine scheduling uses 3\%
warmup; checkpoint 512 is fixed for evaluation.
Routing continues the common linked model with frozen Ground/Relation
teachers, temperature 0.9, and a maximum response length of 256 tokens.
Both loss coefficients equal one, $\lambda_{\rm ref}=\lambda_{\rm KD}=1$. Flat and semantic-uniform allocation use
$\alpha=0$; structural, declared, and semantic-weighted variants use
$\alpha=2$. Teacher-scored positions are measured separately
(Table~\ref{tab:core-seeds}); the response-length limit is common across
selectors.

\paragraph{Controlled repair and data-quality configurations.}
The local Qwen3.6-27B repair study uses greedy decoding and a 512-token limit. Both repair arms receive the original answer and one call per failed candidate. Named feedback additionally gives the first failed category, checked in the order
serialization, typed fields, localization, and depth order, without reference
coordinates or ranks. In the matched-size comparison, each training arm
contains 1,137 records matched by physical-scene counts, with the fresh-base
recipe in Table~\ref{tab:appendix-training-settings}. Individual images
and tokens can differ. The nested-pool subsets and their record counts
are reported separately in Table~\ref{tab:measured-data-scaling}.

\paragraph{Failure-localized repair prompt.}
In the controlled repair study, the named-feedback prompt appends
the previous answer and the failed category to the original task.
For a stored grounding failure, the added instruction is:
\begin{introquestions}\small
\textit{The verifier reports: G: localization does not match the requested
entities. Correct this failure and return a complete response.}
\end{introquestions}
The generic-retry arm instead requests review and correction of the previous answer without naming the failed stage. Both retain the original task, and the feedback does not supply the reference boxes or target depth ranks. Aggregate repair results are reported in
Table~\ref{tab:repair-productivity}.

\begin{table}[!htbp]
\centering\small
\renewcommand{\arraystretch}{1.08}\setlength{\tabcolsep}{5pt}
\caption{\textbf{Point-cardinality sensitivity on the common 286-item VABench subset
with identical image exclusions and saved responses.} $\dagger$: earlier source-filtered \method checkpoint.}
\label{tab:expanded-local}
\begin{tabular}{@{}lrr@{}}\toprule
\rowcolor{anvilAppHead}
\textbf{Model} & \textbf{Any-point (\%)} & \textbf{First-point (\%)} \\
\midrule
Base & 42.31 & 36.71 \\
\rowcolor{anvilAppFocus}
\method\textsuperscript{$\dagger$} & \appbest{81.12} & \appbest{71.33} \\
\bottomrule\end{tabular}
\end{table}
The subset excludes the union of 14 exact/near-image candidates identified
against earlier corpora, identically for every model and without using
correctness. This historical image-exclusion check is distinct from the scene/image
partitioning used for the held-out corpus. On the
full 300 cases, source-filtered and base models return 8.92 and 1.90 points
per response on average, motivating the first-point comparison.

\section{Complete Controlled Comparisons}
\label{app:tradeoffs}
\subsection{Coverage and retention}
\begin{table}[!htbp]
\centering\small
\renewcommand{\arraystretch}{1.08}\setlength{\tabcolsep}{5pt}
\caption{\textbf{Source-composition controls (\%), seed 43 / seed 44.} Updates and
row exposures are matched; realized tokens are not. Bold identifies an arm that leads in both seeds.}
\label{tab:source-controls}\label{tab:source-coverage}
\begin{tabular}{@{}lrrr@{}}\toprule
\rowcolor{anvilAppHead}
\textbf{Benchmark} & \textbf{Replay} & \textbf{Vacancy} & \textbf{Mixed} \\
\midrule
Where2Place & 38.00 / 36.22 & \appbest{69.00 / 69.00} & 68.00 / 66.71 \\
RoboSpatial context & 44.14 / 41.74 & \appbest{63.11 / 62.30} & 62.30 / 59.84 \\
VABench-Point & 79.67 / 80.33 & 77.67 / 76.67 & 80.67 / 78.33 \\
BLINK depth & \appbest{90.32 / 90.32} & 86.29 / 88.71 & 89.52 / 89.52 \\
ERQA & \appbest{50.17 / 49.83} & 46.69 / 46.34 & 48.08 / 48.08 \\
EmbSpatial & \appbest{78.43 / 78.24} & 76.95 / 77.42 & 77.91 / 77.77 \\
\bottomrule\end{tabular}
\end{table}
Mixed supervision improves placement over replay in both runs; vacancy alone
maximizes placement while weakening several retained capabilities. Mixed
training does not dominate all tasks. Increasing only the source-filtered
model's Where2Place response budget from 256 to 512 changes success from
35.16\% to 32.96\% on the same 100 cases.

\subsection{Correction-stage selection}

\begin{table}[!htbp]
\centering\small
\renewcommand{\arraystretch}{1.08}\setlength{\tabcolsep}{5pt}
\caption{\textbf{Observed paired routing contrasts on strict Ground--Relation success.}
Differences are percentage points; 95\% scene-bootstrap intervals condition
on the two recorded training seeds. Per-seed values and teacher-position
counts appear in Table~\ref{tab:core-seeds}.}
\label{tab:observed-routing-contrasts}
\begin{tabular}{@{}lrr@{}}
\toprule
\rowcolor{anvilAppHead}
\textbf{Comparison} & \textbf{$\Delta$ (pp)} & \textbf{95\% CI} \\
\midrule
\rowcolor{anvilAppFocus}
Semantic uniform $-$ flat & \appgain{+2.13} & [0.39, 3.90] \\
Semantic uniform $-$ declared & +0.53 & [-0.61, 1.63] \\
Semantic weighted $-$ uniform & -0.53 & [-1.57, 0.42] \\
\bottomrule
\end{tabular}
\end{table}

VQA retention compares post- and pre-distillation performance on AI2D,
ScienceQA, and TextVQA panels of 309, 202, and 500 examples. A ratio can exceed
100\% when a panel improves.

\begin{table}[!htbp]
\centering\small
\renewcommand{\arraystretch}{1.08}\setlength{\tabcolsep}{5pt}
\caption{\textbf{Per-seed routing observations.} Strict G$\to$S success on 282 images / 50 scenes and sampled VQA retention, reported as seed 17 / seed 23. Teacher positions count scored logits including distributed padding; realized budgets are measured, not matched.}
\label{tab:core-seeds}
\begin{tabular}{@{}lrrr@{}}
\toprule
\rowcolor{anvilAppHead}
\textbf{Method} & \textbf{G$\to$S (\%)} & \textbf{VQA ratio (\%)} & \textbf{Teacher positions} \\
\midrule
Flat declared-teacher & 69.50 / 70.92 & 100.32 / 99.96 & 131,334 / 131,704 \\
Structural route & 71.28 / 70.92 & 100.12 / 99.84 & 221 / 74 \\
Declared route & 71.63 / 71.99 & 100.43 / 99.70 & 56,267 / 56,449 \\
\rowcolor{anvilAppFocus}
Semantic (uniform) & \appbest{72.70 / 71.99} & 99.70 / 99.97 & 47,648 / 48,213 \\
Semantic (weighted) & 71.99 / 71.63 & 99.77 / 99.85 & 48,066 / 48,130 \\
\bottomrule
\end{tabular}
\end{table}

Semantic-uniform exceeds flat in both seeds, with a paired gain of
2.13 percentage points (95\% CI: [0.39, 3.90]).
Both use uniform token weights, supporting selective teacher--span
allocation under common training settings.
Table~\ref{tab:observed-routing-contrasts} additionally compares
routing configurations and token-weighting choices;
stage normalization is shared across configurations.

\subsection{Repair yield and downstream learning}

\begin{table}[!htbp]
\centering\small
\renewcommand{\arraystretch}{1.08}\setlength{\tabcolsep}{5pt}
\caption{\textbf{Observed filtering counts in the shared 2,011-candidate pool.}
Each percentage uses all initial candidates as its denominator. Source
correctness is evaluated against source references, not human precision.}
\label{tab:measured-pipeline}
\begin{tabular}{@{}lrr@{}}
\toprule
\rowcolor{anvilAppHead}
\textbf{Condition} & \textbf{Candidates retained} & \textbf{Fraction of initial pool (\%)} \\
\midrule
Initial candidates & 2,011 & 100.00 \\
Parseable & 1,998 & 99.35 \\
Pass typed checks & 1,965 & 97.71 \\
\rowcolor{anvilAppFocus}
Source-correct & 1,126 & 55.99 \\
\bottomrule
\end{tabular}
\end{table}

The filtering and repair denominators differ. Structural retention uses all
$2{,}011$ initial candidates; conditional source correctness after typed checks
is $1{,}126/1{,}965=57.30\%$. The $885$ initial failures form the repair pool.
Accepting only source-correct outputs makes conditional acceptance precision
100\% by construction, so that number cannot independently calibrate a
verifier.

\begin{table}[!htbp]
\centering\small
\renewcommand{\arraystretch}{1.08}\setlength{\tabcolsep}{5pt}
\caption{\textbf{Observed repair productivity.} Both arms receive the same 885 rejected
candidates and make one call per candidate. GPU-hours measure allocated
accelerator time; source-correct recoveries supply the yield numerator.}
\label{tab:repair-productivity}
\begin{tabular}{@{}lrrrr@{}}
\toprule
\rowcolor{anvilAppHead}
\textbf{Repair arm} & \textbf{Correct} & \textbf{Yield (\%)} & \textbf{GPU-hours} & \textbf{Correct / GPU-hour} \\
\midrule
Generic retry & 11 & 1.24 & 2.7664 & 3.98 \\
\rowcolor{anvilAppFocus}
Named feedback & \appbest{111} & \appbest{12.54} & 2.8186 & \appbest{39.38} \\
\bottomrule
\end{tabular}
\end{table}

Named feedback recovers 111 correct records versus 11 for generic retry
at similar GPU cost. Matched-size feedback training reaches 59.22\%
strict success versus 56.38\% for raw data, a difference of
2.84 points [-1.41, 6.88]. Repair yield improves, but the downstream
learning interval includes zero and nested-pool performance is
nonmonotonic (Table~\ref{tab:measured-data-scaling}).

\begin{table}[!htbp]
\centering\small
\renewcommand{\arraystretch}{1.08}\setlength{\tabcolsep}{5pt}
\caption{\textbf{Source-correct repairs by initial failure category.} The categories
partition the rejected pool according to the first failed check; each row
reports recovery counts and within-category percentages in both arms.}
\label{tab:repair-by-failure}
\begin{tabular}{@{}lrrr@{}}
\toprule
\rowcolor{anvilAppHead}
\textbf{Initial failure} & \textbf{Candidates} & \textbf{Generic retry} & \textbf{Named feedback} \\
\midrule
Serialization & 13 & 0 (0.00\%) & 1 (7.69\%) \\
Typed fields & 33 & 2 (6.06\%) & 3 (9.09\%) \\
Grounding & 761 & 7 (0.92\%) & 42 (5.52\%) \\
\rowcolor{anvilAppFocus}
Depth order & 78 & 2 (2.56\%) & \appbest{65 (83.33\%)} \\
\midrule
\textbf{Total} & 885 & 11 (1.24\%) & \appbest{111 (12.54\%)} \\
\bottomrule
\end{tabular}
\end{table}

Table~\ref{tab:repair-by-failure} makes this concentration explicit: named
feedback recovers $65/78$ depth-order failures but only $42/761$ grounding
failures. The overall multiplier therefore depends on the composition and
information content of the rejected pool.

\paragraph{A retained grounding failure.}
The repair logs also preserve an unsuccessful case with two picture
instances. The initial response passes syntax and typed checks but
fails the source-based grounding criterion, while its depth order
passes. Both generic retry and named grounding feedback return the
same boxes as the initial response. This case complements the
successful relation repair in Figure~\ref{fig:placement-showcase}:
localizing a failure can guide another attempt without guaranteeing
that the model will correct it. The selected cases illustrate failure
modes; their frequency is assessed by the corresponding quantitative
studies.

\paragraph{Cost accounting.}
With 885 calls per arm, the per-call yield ratio is $111/11=10.09$.
Allocated GPU time gives $(111/2.8186)/(11/2.7664)=9.90$.
These production costs exclude subsequent student training.

\begin{table}[!htbp]
\centering\small
\renewcommand{\arraystretch}{1.08}\setlength{\tabcolsep}{5pt}
\caption{\textbf{Nested source coverage at fixed updates.} Scene-stratified
subsets share fresh-base initialization, 256 updates, 1,024 row exposures,
and seed 17. Strict G$\to$S success (\%) uses 282 images / 50 scenes;
brackets give 95\% scene-bootstrap intervals. Tokens are measured.}
\label{tab:measured-data-scaling}
\begin{tabular}{@{}lrrr@{}}
\toprule
\rowcolor{anvilAppHead}
\textbf{Pool fraction} & \textbf{Unique records} & \textbf{Input tokens} & \textbf{Success (\%)} \\
\midrule
25\% & 284 & 1,524,912 & 59.93 [49.47, 69.73] \\
50\% & 568 & 1,532,961 & 58.87 [48.28, 68.77] \\
100\% & 1,137 & 1,535,783 & 59.22 [50.00, 68.42] \\
\bottomrule
\end{tabular}
\end{table}

At fixed updates and exposures, pool size changes the available unique
records; realized tokens remain unmatched. The seed-17 results establish
neither monotonic data scaling nor a reliable matched-size learning gain.

\begin{figure}[!htb]
\centering
\includegraphics[width=\linewidth]{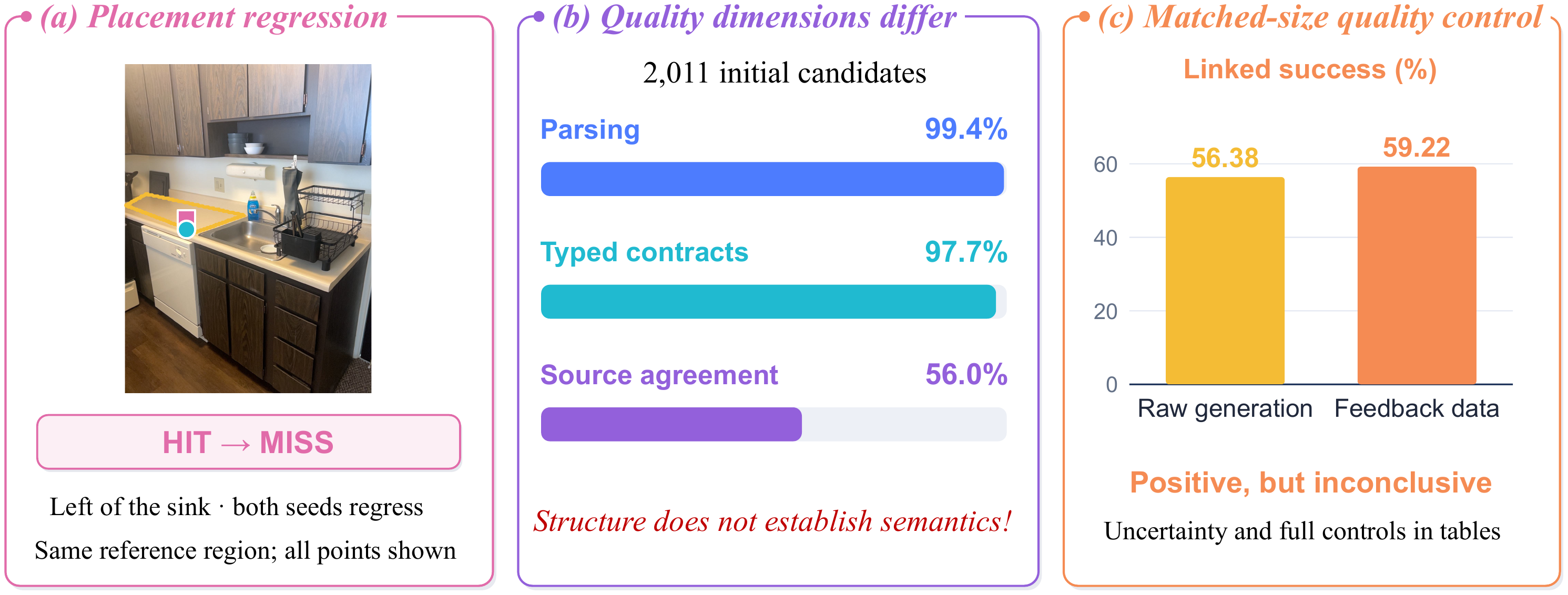}
\caption{\textbf{Failure cases and limits of quality-based learning.}
(a) A saved placement example regresses after adaptation in both seeds;
selected cases do not estimate failure frequency.
(b) Parsing, typed validity, and source correctness all use the same 2,011
initial candidates as the denominator. Conditional source correctness among
the 1,965 typed-valid records is reported separately in the text.
(c) Matched-size feedback training improves the point estimate by 2.84 points,
but the paired interval $[-1.41,6.88]$ includes zero.}
\label{fig:appendix-failure-quality}
\end{figure}

\subsection{Construction and rechecking cost}
A single CPU pass reads 598,892 lines: 598,891 parse and one is malformed.
Parsed records have no violations of the covered structural rules. The pass
includes reading, parsing, hashing, and rule evaluation, but excludes image
loading, generation, and semantic review. Conformance of retained records
does not estimate raw-generation yield or corpus-wide semantic precision.
Aggregate throughput is approximately 45.2K records/s. This timing
study covers eight historical storage types and measures structural
rechecking only. The current data flywheel generates and verifies eleven
formats; model-based semantic verification is outside this CPU timing.

\begin{table}[!htbp]
\centering\small
\renewcommand{\arraystretch}{1.08}\setlength{\tabcolsep}{5pt}
\caption{\textbf{Stored-record rechecking: single-run CPU wall time, no model calls.}}
\label{tab:flywheel-efficiency}
\begin{tabular}{@{}lrrr@{}}\toprule
\rowcolor{anvilAppHead}
\textbf{Shard} & \textbf{Parsed records} & \textbf{Malformed lines} & \textbf{Time (s)} \\
\midrule
0 & 167,459 & 0 & 3.81 \\
1 & 163,286 & 0 & 3.61 \\
2 & 167,682 & 0 & 3.65 \\
3 & 100,464 & 1 & 2.18 \\
\midrule \rowcolor{anvilAppNeutral}
\textbf{Total} & 598,891 & 1 & 13.26 \\
\bottomrule\end{tabular}
\end{table}

\section{Task Structure and Evidence Dependence}
\label{app:scale-evidence}
\paragraph{Linked task construction.}
Eight frames from each of 1,201 ScanNet training scans give 9,608 candidates.
After RGB-D extraction, eight CPU workers construct 2,463 linked records
from 447 scenes in 325.25 seconds using registered depth and instance geometry.
Other frames lack qualifying depth-separated pairs. Visible names identify
856 pairs; interior clicks disambiguate 1,607. Evaluation uses 282 images
from 50 held-out scenes. Linked and atomic-mixture training use 3.07M and
2.79M tokens at equal updates.
Closing saved stage delimiters preserves JSON contents, geometry, labels,
ordering, and finish reasons. Atomic-mixture success rises from 1.77\%
to 44.68\%; linked success remains 69.86\%, leaving a 25.18-point gap.
This linked-interface advantage coexists with weaker atomic-task performance
(Table~\ref{tab:linked-source-training}).

\begin{table}[!htbp]
\centering\small
\renewcommand{\arraystretch}{1.08}\setlength{\tabcolsep}{5pt}
\caption{\textbf{Atomic and linked source training.} All four arms use
512 updates and 2,048 row exposures; evaluation uses 282 images from
50 scenes. Detection and Depth use atomic-task prompts, whereas the linked
columns use Ground--Relation prompts. Scores are percentages.
\emph{Tags closed} changes only saved closing delimiters; training tokens
are recorded rather than matched. Bold marks the highest score in each column.}
\label{tab:linked-source-training}
\begin{tabular}{@{}lrrrr@{}}
\toprule
\rowcolor{anvilAppHead}
\textbf{Training view} & \multicolumn{2}{c}{\textbf{Atomic tasks}}
              & \multicolumn{2}{c}{\textbf{Linked tasks}} \\
\cmidrule(lr){2-3}\cmidrule(lr){4-5}
 & \textbf{Detection} & \textbf{Depth} & \textbf{Strict G$\to$S} & \textbf{Tags closed} \\
\midrule
Ground only & \appbest{84.75} & 0.00 & 18.79 & 50.71 \\
Relation only & 20.60 & \appbest{94.68} & 0.00 & 0.71 \\
Atomic mix & 84.04 & 94.33 & 1.77 & 44.68 \\
\rowcolor{anvilAppFocus}
Linked & 25.71 & 0.00 & \appbest{69.86} & \appbest{69.86} \\
\bottomrule
\end{tabular}
\end{table}

\paragraph{Visual and upstream interventions.}

\begin{table}[!htbp]
\centering\small
\renewcommand{\arraystretch}{1.08}\setlength{\tabcolsep}{5pt}
\caption{\textbf{Visual-input interventions.} Full-split scores (\%) with
questions and targets fixed. Parentheses give sample counts; failures remain
in the denominator. $\dagger$: earlier source-filtered \method checkpoint, not the final model.}
\label{tab:visual-controls}
\begin{tabular}{@{}llrrr@{}}
\toprule
\rowcolor{anvilAppHead}
\textbf{Model} & \textbf{Benchmark} & \textbf{Original} & \textbf{Blank} & \textbf{Shuffled} \\
\midrule
\rowcolor{anvilAppNeutral}
Base & \benchg{} (1,463) & 56.87 & 3.69 & 3.69 \\
Base & BLINK (124) & 90.32 & 48.39 & 47.58 \\
Base & RoboSpatial (122) & 44.12 & 2.79 & 1.80 \\
\addlinespace[3pt]
\rowcolor{anvilAppFocus}
\method\textsuperscript{$\dagger$} & \benchg{} (1,463) & 57.35 & 7.66 & 3.49 \\
\method\textsuperscript{$\dagger$} & BLINK (124) & 89.52 & 51.61 & 48.39 \\
\method\textsuperscript{$\dagger$} & RoboSpatial (122) & 33.60 & 0.30 & 4.50 \\
\bottomrule
\end{tabular}
\end{table}

\begin{table}[!htbp]
\centering\small
\renewcommand{\arraystretch}{1.08}\setlength{\tabcolsep}{5pt}
\caption{\textbf{Upstream localization interventions.} Depth accuracy (\%)
on 282 images from 50 scenes with fixed depth targets. Differences are
specialist minus joint, with 95\% paired scene-bootstrap intervals
(5,000 draws). Both models use 512 source-SFT updates.}
\label{tab:source-context}
\begin{tabular}{@{}lrrr@{}}
\toprule
\rowcolor{anvilAppHead}
\textbf{Upstream boxes} & \textbf{Relation specialist} & \textbf{Joint} & \textbf{Specialist $-$ joint} \\
\midrule
Reference & 94.68 & 94.33 & +0.35 [-1.79, +2.43] \\
Student-produced & 93.62 & 92.20 & +1.42 [-1.54, +4.38] \\
Shifted & 84.04 & 82.62 & +1.42 [-2.13, +4.91] \\
\bottomrule
\end{tabular}
\end{table}

Blank inputs are constant-gray images; shuffled images come from another
scene, with instructions and targets fixed. Upstream interventions replace
reference boxes with predicted or shifted boxes while preserving images
and depth targets. These controls establish dependence on supplied evidence,
not an internal reasoning mechanism.
Reference boxes provide clean context; predicted boxes expose upstream model
errors; shifted boxes introduce a controlled perturbation. All three
specialist--joint intervals include zero, so their point estimates do not
establish a specialist advantage.

\paragraph{Action-order verification.}

\begin{table}[!htbp]
\centering\small
\renewcommand{\arraystretch}{1.08}\setlength{\tabcolsep}{5pt}
\caption{\textbf{State-conditioned plan verification.} Accuracy (\%) on
1,012 balanced judgments from 506 ALFRED trajectories and 92 scenes.
Each valid plan is paired with an invalid ordering of the same actions;
95\% intervals resample scenes. State and rules are supplied. $\dagger$: earlier source-filtered \method checkpoint, not the final model. Bold marks the highest point estimate in each accuracy column.}
\label{tab:alfred-plan-verification}
\begin{tabular}{@{}lrrrr@{}}
\toprule
\rowcolor{anvilAppHead}
\textbf{Model} & \textbf{Valid} & \textbf{Invalid} & \textbf{Overall $\uparrow$} & \textbf{95\% CI} \\
\midrule
Qwen3.5-9B & 40.51 & \appbest{91.11} & 65.81 & [62.07, 70.60] \\
\method\textsuperscript{$\dagger$} & \appbest{59.09} & 77.67 & 68.38 & [65.42, 72.57] \\
\rowcolor{anvilAppFocus}
Action specialist & 55.53 & 89.53 & \appbest{72.53} & [68.22, 78.41] \\
Joint SFT & 53.56 & 88.54 & 71.05 & [66.86, 76.65] \\
\bottomrule
\end{tabular}
\end{table}

Each of 506 ALFRED trajectories supplies a valid plan and an invalid
permutation with identical actions and length. Models receive the initial
image, complete Boolean state, goal, and source action rules. This pairing
removes action occurrence as a sufficient label cue; class-wise scores
expose the remaining invalid-answer bias. The Action specialist is separate
from the final MOPD model. Because state and rules are supplied, the task
measures state-conditioned validity rather than visual-only planning or
standard ALFRED execution.

\paragraph{Qualitative failures and quality controls.}
In Figure~\ref{fig:appendix-failure-quality}, the saved
``left of the sink'' prediction regresses after adaptation in both seeds.
The remaining panels separately measure structural validity, source
correctness, and downstream learning.

\section{Source-native Specialist Preparation}
\label{app:native-preparation}

Table~\ref{tab:native-preparation} compares terminal specialists with joint SFT
after 1,024 LoRA updates from Qwen3.5-9B (seed 17), before MOPD.
It covers all examples in the completed source-grounded forms
(Appendix~\ref{app:scene-sources}), with matched inputs, greedy decoding,
and scoring. Malformed or truncated outputs remain in the denominator.
The pretrained baseline receives the same task inputs and generation budgets,
with each checkpoint using its own chat template.

\begin{table}[htbp]
\appTableSetup
\caption{\textbf{Source-Native Preparation and Pretrained Baseline}:
Percentages, higher is better; Relation specialist for depth, Act otherwise.
RoboBrain denotes RoboBrain2.5-8B-NV~\citep{tan2026robobrain25}.
The ALFRED rows share 3,327 examples. Reference
agreement and output validity do not measure execution or grasp success.}
\label{tab:native-preparation}
\begin{tabular}{llrrrr}
\toprule
Source & Metric $\uparrow$ & $N$ & Specialist & Joint & RoboBrain \\
\midrule
PRISM & Depth agreement & 3,601 & 91.67 & 88.14 & 0.00 \\
ALFRED & Exact continuation & 3,327 & 25.91 & 18.18 & 0.00 \\
ALFRED & Next-action agreement & 3,327 & 64.44 & 56.03 & 0.00 \\
ShareRobot-Bench & Region IoU $\geq 0.5$ & 193 & 5.70 & 6.74 & 23.32 \\
PRISM & Valid gripper output & 580 & 0.00 & 0.00 & 0.00 \\
\bottomrule
\end{tabular}
\end{table}

Depth and planning gains are task-specific. RoboBrain achieves 45/193 on
affordance IoU50, above both preparation models, but all three fail the
gripper-output contract. For Action, 653/3,327 baseline outputs are structurally
valid; none matches the reference next-action label. Its 3,601 depth replies
are JSON but fail the native contract. Strict scoring combines spatial
prediction with \mbox{interface and label adherence}.

\section{Limitations and Future Work}
\label{app:limitations}\label{app:completion-tests}

\paragraph{From predefined interfaces to open-world spatial supervision.}
The current framework organizes supervision through eleven
Ground--Relation--Act forms with explicit entities and dependencies.
Open-world environments introduce spatial concepts and interaction
requirements beyond this inventory. Extensible interfaces could incorporate
new relations, affordances, and temporal dependencies while retaining a
common basis for generation, verification, and learning. The challenge is
to expand this vocabulary without losing the scene evidence supporting
each target.

\paragraph{From specialist consolidation to compositional spatial competence.}
Anvil studies knowledge transfer within specified spatial tasks and
stage dependencies. Future work could extend specialist-to-generalist
learning to unfamiliar combinations of goals, environments, and embodiments.
The question is whether grounding, relational reasoning, and action
knowledge can compose into reusable spatial expertise, supporting new
tasks with less task-specific supervision.

\paragraph{From knowledge recirculation to learning through interaction.}
The present framework draws on curated sources and explicit verification
signals. Spatial uncertainty and encountered failures could guide what an
agent observes and attempts next, with physical interaction outcomes
grounding the returned supervision. Connecting experience collection to
specialist refinement and continued generalist learning remains a future
direction: stronger spatial competence should support both better decisions
and more informative data acquisition.

\end{document}